\documentclass[10pt,twocolumn,letterpaper]{article}

\usepackage[pagenumbers]{cvpr} %

\usepackage{graphicx}
\usepackage{amsmath}
\usepackage{amssymb}
\usepackage{booktabs}
\usepackage{adjustbox}

\usepackage[accsupp]{axessibility}  %

\usepackage[pagebackref,breaklinks,colorlinks]{hyperref}

\usepackage[dvipsnames,table,xcdraw]{xcolor}
\makeatletter
\@namedef{ver@everyshi.sty}{}
\makeatother
\usepackage{tikz}
\usepackage{color}
\usepackage{placeins}

\usepackage{array}
\usepackage{multirow}
\usepackage[flushleft]{threeparttable}
\usepackage{comment}
\usepackage{algorithm}
\usepackage{algpseudocode}
\usepackage{booktabs}

\usepackage{placeins}
\usepackage{float}

\usepackage{amssymb}%
\usepackage{pifont}%

\usepackage{fancyvrb}

\usepackage{graphicx}
\usepackage{enumitem}
\usepackage{wrapfig}
\usepackage{lipsum}

\usepackage{soul}
\usepackage[bb=dsserif]{mathalpha}
\usepackage{bm}

\usepackage{makecell}

\usepackage[capitalize]{cleveref}
\crefname{section}{Sec.}{Secs.}
\Crefname{section}{Section}{Sections}
\Crefname{table}{Table}{Tables}
\crefname{table}{Tab.}{Tabs.}

\newcommand{\cmark}{\text{\ding{51}}}%
\newcommand{\xmark}{\text{\ding{55}}}%

\newcommand{\ol}[3]{\begin{#1}[leftmargin=*,topsep=0pt]\setlength{\itemsep}{#2mm}#3\end{#1}}

\newcommand{\tablestyle}[2]{\setlength{\tabcolsep}{#1}\renewcommand{\arraystretch}{#2}\centering\footnotesize}
\newlength\savewidth\newcommand\shline{\noalign{\global\savewidth\arrayrulewidth
  \global\arrayrulewidth 1pt}\hline\noalign{\global\arrayrulewidth\savewidth}}

\def\tableonefont#1#{
    \fontsize{8}{12}
    #1
    \selectfont
}

\def\fontsmall#1#{
    \fontsize{8}{12}
    #1
    \selectfont
}

\def\tablefontsmall#1#{
    \fontsize{8}{10}
    #1
    \selectfont
}

\def\tablefont#1#{
    \fontsize{9}{12}
    #1
    \selectfont
}

\def\gray#1{\textcolor{gray}{#1}}
\definecolor{textpink}{RGB}{255,66,161}

\let\vec\mathbf

\usepackage[pagebackref,breaklinks,colorlinks]{hyperref}

\def\cvprPaperID{4116} 
\def\confName{CVPR}
\def\confYear{2023}

\begin{document}

\title{
\mbox{
\hspace{-20pt}%
Bootstrapping Objectness from Videos by Relaxed Common Fate and Visual Grouping
}
}

\author{
Long Lian$^\text{1}$ \\
$^\text{1}$UC Berkeley \\
{\tt\small longlian@berkeley.edu}
\and
Zhirong Wu$^\text{2}$  \\
$^\text{2}$Microsoft Research Asia \\
{\tt\small 
wuzhiron@microsoft.com}
\and
Stella X. Yu$^\text{1,3}$ \\
$^\text{3}$University of Michigan \\
{\tt\small 
stellayu@umich.edu}
}

\maketitle

\begin{abstract}
We study learning object segmentation from unlabeled videos. Humans can easily segment moving objects without knowing what they are. The Gestalt law of common fate, i.e., what move at the same speed belong together, has inspired unsupervised object discovery based on motion segmentation. However, common fate is not a reliable indicator of objectness: Parts of an articulated / deformable object may not move at the same speed, whereas shadows / reflections of an object always move with it but are not part of it.

Our insight is to bootstrap objectness by first learning image features from relaxed common fate and then refining them based on visual appearance grouping within the image itself and across images statistically. Specifically, we learn an image segmenter first in the loop of approximating optical flow with constant segment flow plus small within-segment residual flow, and then by refining it for more coherent appearance and statistical figure-ground relevance.

On unsupervised video object segmentation, using only ResNet and convolutional heads, our model surpasses the state-of-the-art by absolute gains of $7/9/5\%$ on DAVIS16 / STv2 / FBMS59 respectively, demonstrating the effectiveness of our ideas. Our code is publicly available.
\end{abstract}
\def\im#1#2{\includegraphics[width=0.185\linewidth,height=0.185\linewidth]{figTeaser/#1/#2}}
\def\imres#1#2{
\rotatebox{90}{\textcolor{red}{#2}}&
\im{Images}{#1.jpg}&
\im{Flows}{#1.jpg}&
\im{AMD}{#1.png}&
\im{OCLR}{#1.png}&
\im{Ours}{#1.png}\\[-2pt]
}
\def\figCommonFateTeaser#1{
\begin{figure}[#1]
\centering
\setlength{\tabcolsep}{1pt}
\includegraphics[width=\linewidth]{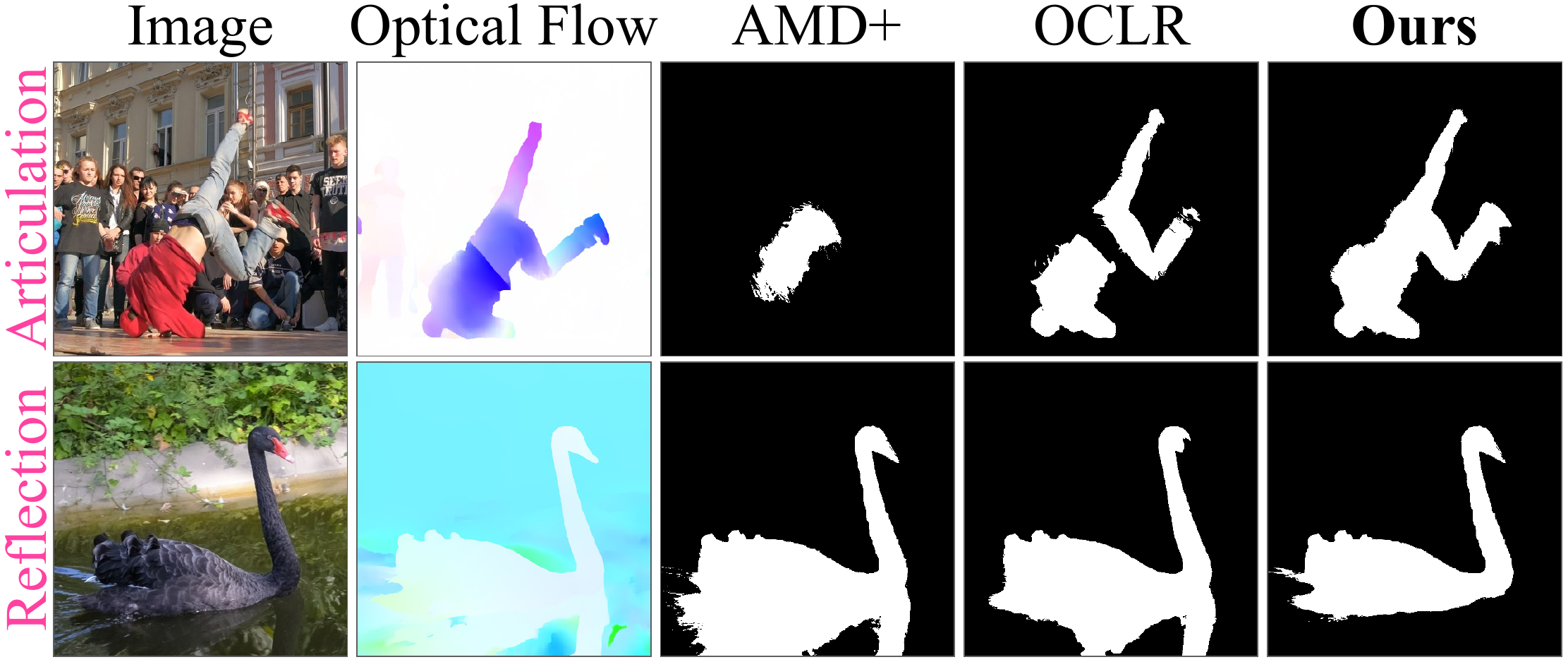}
\vspace{-16pt}
\caption{\textbf{We study how to discover objectness from unlabeled videos based on common motion and appearance.}   AMD\cite{liu2021emergence} and OCLR\cite{xie2022segmenting} rely on {\it common fate}, \ie, what move at the same speed belong together, which is not always a reliable indicator of objectness. {\bf Top}: \textcolor{textpink}{Articulation} of a human body means that object parts may not move at the same speed; common fate thus leads to {\it partial objectness}. {\bf Bottom}: \textcolor{textpink}{Reflection} of a swan in water always moves with it but is not part of it; common fate thus leads to {\it excessive objectness}.  Our method discovers full objectness by relaxed common fate and visual grouping. 
AMD+ refers to AMD with RAFT flows as motion supervision for fair comparison.
\vspace{-15pt}
}
\label{fig:commonFateTeaser}
\end{figure}
}

\def\figAdvantage#1{
\begin{figure}[#1]
\centering
\setlength{\tabcolsep}{1pt}
\begin{tabular}{@{}lcccc@{}}
\toprule
Unsupervised object segmentation & 
MG & AMD & GWM & \textbf{Ours}\\
\midrule
{Sources of supervision} & M & M$^*$ & M & M+A \\
{Segment stationary objects?} & \color{Red}{\xmark} & \color{Green}{\cmark} & \color{Green}{\cmark} & \color{Green}{\cmark} \\
{Handle articulated objects?} & - & \color{Red}{\xmark} & \color{Red}{\xmark} & \color{Green}{\cmark} \\
{Label-free hyperparameter tuning?} & \color{Red}{\xmark} & \color{Red}{\xmark} & \color{Red}{\xmark} & \color{Green}{\cmark} \\
\bottomrule
\end{tabular}\\
\vspace{-1pt}
\includegraphics[width=1.0\linewidth]{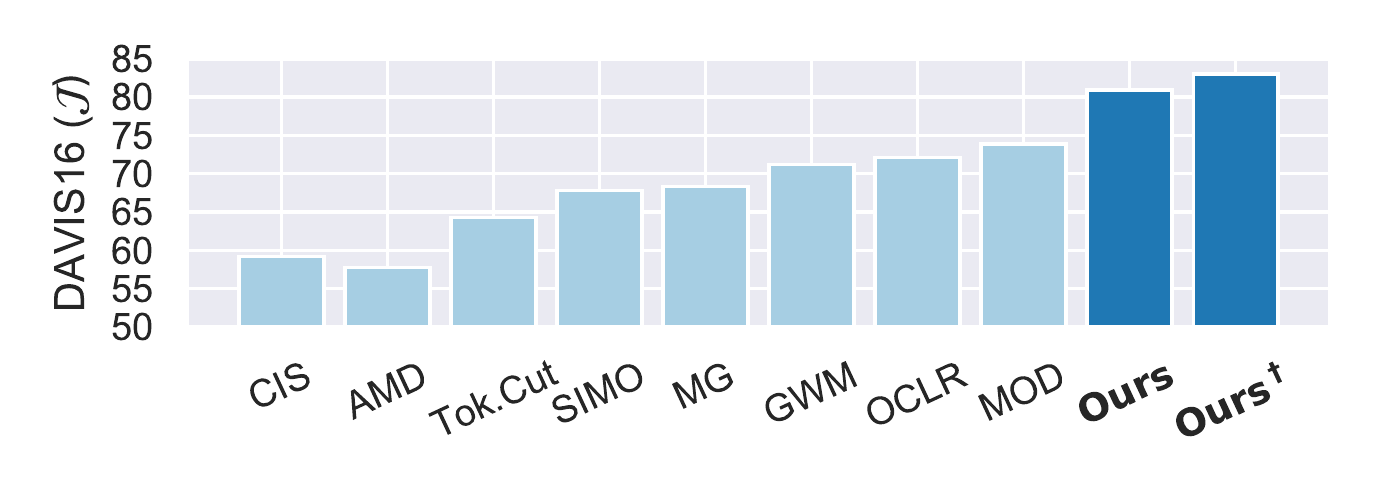}
\vspace{-30pt}
\caption{Advantages 
over leading unsupervised object segmentation methods MG \cite{yang2021self}/AMD \cite{liu2021emergence}/GWM \cite{choudhury2022guess}:
{\bf 1)} With motion supervision instead of motion input, we can segment stationary objects.
{\bf 2)} With both motion (M) and appearance (A) as supervision, we can discover full objectness from noisy motion cues. 
M$^*$ refers to implicit motion via image warping.
{\bf 3)} By modeling relative motion within an object, we can handle articulated objects. 
{\bf 4)} By comparing motion-based segmentation with appearance-based segmentation, we can tune hyperparameters without labels.
Our performance gain is substantial, more with 
post-processing ($\dagger$).\\[-20pt]
}
\label{fig:advantage}
\end{figure}
}

\section{Introduction}
\label{introduction}

Object segmentation from videos is useful to many vision and robotics tasks \cite{perazzi2016benchmark,li2013video,brox2010freiburg,ochs2013segmentation}.  However, most methods rely on pixel-wise human annotations \cite{lu2019see,mahadevan2020making,zhen2020learning,li2018unsupervised,schmidt2022d2conv3d,ren2021reciprocal,zhou2020motion,ji2021full,cheng2021rethinking,cheng2022xmem,liu2022learning,miao2022region}, limiting their practical applications. 

We focus on learning object segmentation from entirely {\it unlabeled} videos (Fig.~\ref{fig:commonFateTeaser}). The Gestalt law of {\it common fate}, i.e., {\it what move at the same speed belong together}, has inspired a large body of unsupervised object discovery based on motion segmentation \cite{yang2019unsupervised,yang2021self,meunier2022driven,liu2021emergence,lamdouar2021segmenting,xie2022segmenting,choudhury2022guess}.

\figCommonFateTeaser{tp}

There are three main types of {\it unsupervised} video object segmentation (UVOS) methods.  {\bf 1) Motion segmentation} methods \cite{yang2021self, meunier2022driven,lamdouar2021segmenting,xie2022segmenting} use motion signals from a pretrained optical flow estimator to segment an image into foreground objects and background (Fig. \ref{fig:commonFateTeaser}). OCLR \cite{xie2022segmenting} achieves state-of-the-art performance by first synthesizing a dataset with arbitrary objects moving and then training a motion segmentation model with known object masks.
{\bf 2) Motion-guided image segmentation} methods such as GWM \cite{choudhury2022guess} use motion segmentation loss to guide appearance-based segmentation. Motion between video frames is only required during training, not during testing.
{\bf 3) Joint appearance segmentation and motion estimation} methods such as AMD \cite{liu2021emergence} learn motion and segmentation simultaneously in a self-supervised fashion by reconstructing the next frame based on how segments of the current frame move.

However, while {\it common fate} is effective at binding parts of heterogeneous appearances into one whole moving object, it is not a reliable indicator of objectness (Fig.~\ref{fig:commonFateTeaser}).
\ol{enumerate}{-1}{
\item {\bf Articulation}: Parts of an articulated or deformable object may not move at the same speed; common fate thus leads to {\it partial} objectness containing the major moving part only.  In Fig.\ref{fig:commonFateTeaser} top, AMD+ discovers only the middle torso of the street dancer since it moves the most, whereas OCLR misses the exposed belly which is very different from the red hoodie and the gray jogger.

\item {\bf Reflection}: Shadows or reflections of an object always move with the object but are not part of the object; common fate thus leads to {\it excessive} objectness that covers more than the object.  In Fig.\ref{fig:commonFateTeaser} bottom, AMD+ or OCLR cannot separate the swan {\it from} its reflection in water.
}

\figAdvantage{tp}

We have two insights to bootstrap full objectness from common fate in unlabeled videos.  {\bf 1)} To detect an articulated object, we allow various parts of the same object to assume different speeds that deviate slightly from the object's overall speed.    {\bf 2)}  To detect an object from its reflections, we rely on visual appearance grouping within the image itself and statistical figure-ground relevance. For example, swans tend to have distinctive appearances from the water around them, and reflections may be absent in some swan images.

Specifically, we learn unsupervised object segmentation in two stages: Stage 1 learns to discover objects from motion supervision with relaxed common fate, whereas Stage 2 refines the segmentation model based on image appearance.

{\bf At Stage 1}, we discover objectness by computing the optical flow and learning an image segmenter in the loop of approximating the optical flow with {\it constant segment flow} plus {\it small within-segment residual flow}, relaxing {\it common fate} from the strict same-speed assumption.
{\bf At Stage 2}, we refine our model by image appearance based on low-level visual coherence within the image itself and usual figure-ground distinction learned statistically across images.

Existing UVOS methods have hyperparameters that significantly impact the quality of segmentation.  For example, the number of segmentation channels is a critical parameter for AMD \cite{liu2021emergence}, and 
it is usually chosen according to an annotated validation set in the downstream task, defeating the claim of {\it unsupervised} objectness discovery.

We propose \textbf{unsupervised hyperparameter tuning} that does not require any annotations.
We examine how well our motion-based segmentation aligns with appearance-based affinity on DINO\cite{caron2021emerging} features self-supervisedly learned on ImageNet\cite{ILSVRC15}, which is known to capture semantic objectness.  Our idea is also \textit{model-agnostic} and applicable to other UVOS methods.

Built on the novel concept of \underline{R}elaxed 
\underline{C}ommon 
\underline{F}ate (RCF),
our method
has several advantages over leading UVOS methods (Fig.~\ref{fig:advantage}):
It is the only one that uses both motion and appearance to supervise learning; it can segment stationary and articulated objects in single images, and it can tune hyperparameters without external annotations.

On UVOS benchmarks, using only standard ResNet\cite{he2016deep} backbone and convolutional heads,
our RCF surpasses the state-of-the-art by absolute gains of $7.0\%/9.1\%/4.5\%$ ($6.3\%/12.0\%/5.8\%$) without (with) post-processing on  DAVIS16\cite{perazzi2016benchmark} / STv2\cite{li2013video} / FBMS59\cite{ochs2013segmentation} respectively, validating the effectiveness of our ideas.

\def\figMotionSup#1{
\begin{figure*}[#1]
\centering
\vspace{-15pt}
\centerline{\includegraphics[width=1.0\linewidth]{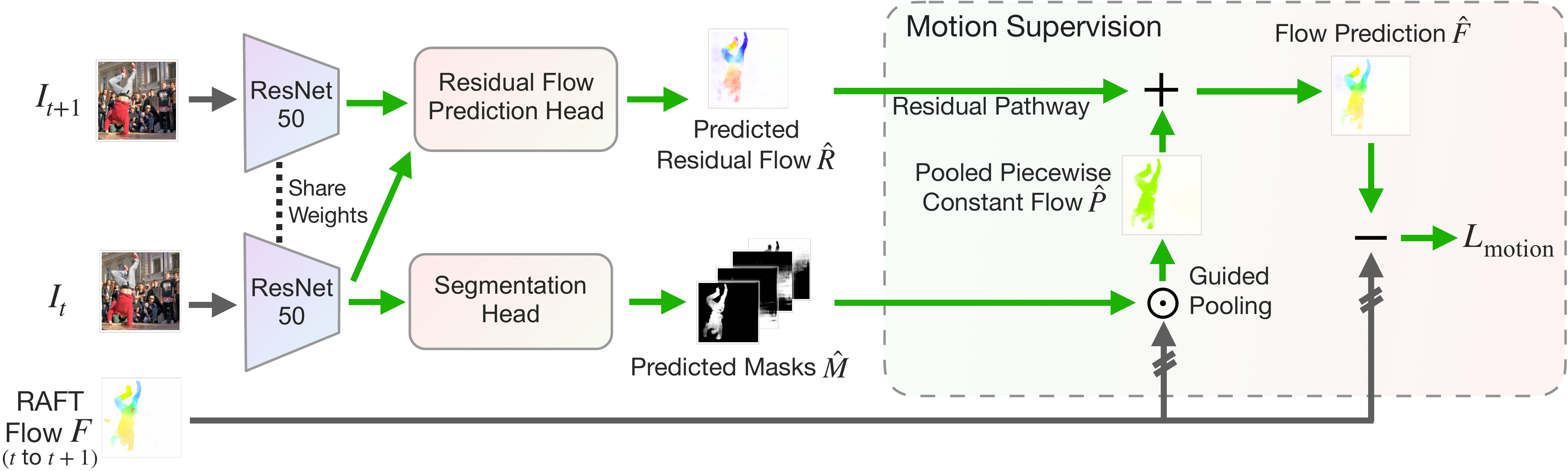}}
\vspace{-10pt}
\caption{
\textbf{Our object discovery stage uses motion as supervision and follows the principle of relaxed common fate}, in which training signals are obtained by reconstructing the reference RAFT flow with the sum from the two pathways: \textbf{1)} a piecewise constant flow pathway, which is created from pooling the RAFT flow with the predicted masks in order to model object-level motion; \textbf{2)} a predicted pixel-wise residual flow pathway, which models intra-object motion for articulated and deformable objects. Green arrows indicate gradient backprop.
}
\label{fig:motion-sup}
\end{figure*}
}

\def\figAppearanceSup#1{
\begin{figure*}[#1]
\centering
\vspace{-20pt}
\centerline{\includegraphics[width=1.0\linewidth]{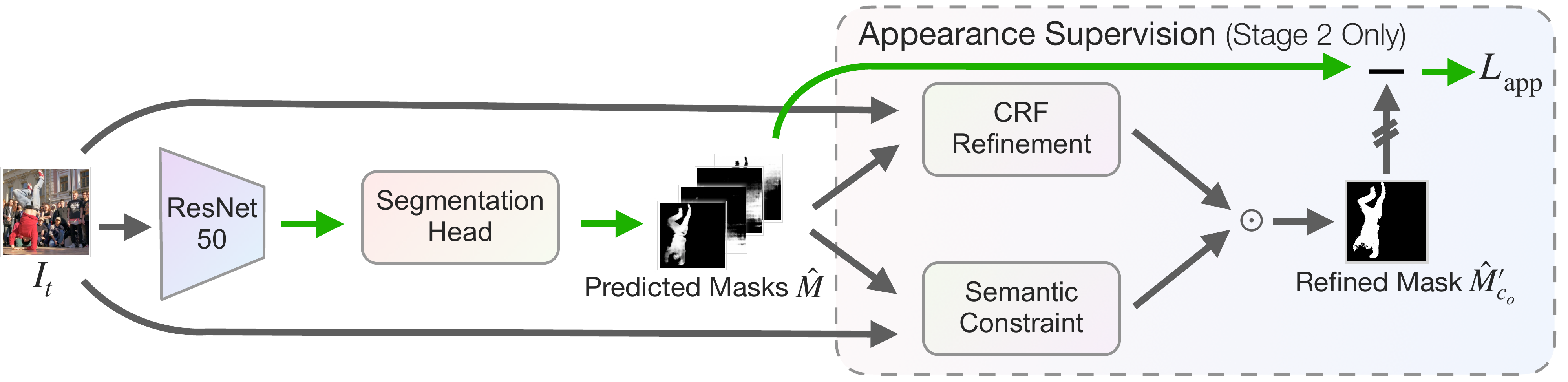}}
\vspace{-5pt}
\caption{
\textbf{Our appearance refinement stage} corrects misconceptions from motion supervision. The predicted mask is supervised by a refined mask based on both the CRF that enforces low-level appearance consistency (\eg, color and texture) and the semantic constraint that enforces high-level semantic consistency. External frozen image features used to enforce the semantic constraint are omitted for clarity. 
}

\label{fig:appearance-sup}
\end{figure*}
}

\def\figRefineExample#1{
\begin{figure}[#1]
\vspace{-5pt}
\centerline{\includegraphics[width=0.925\linewidth]{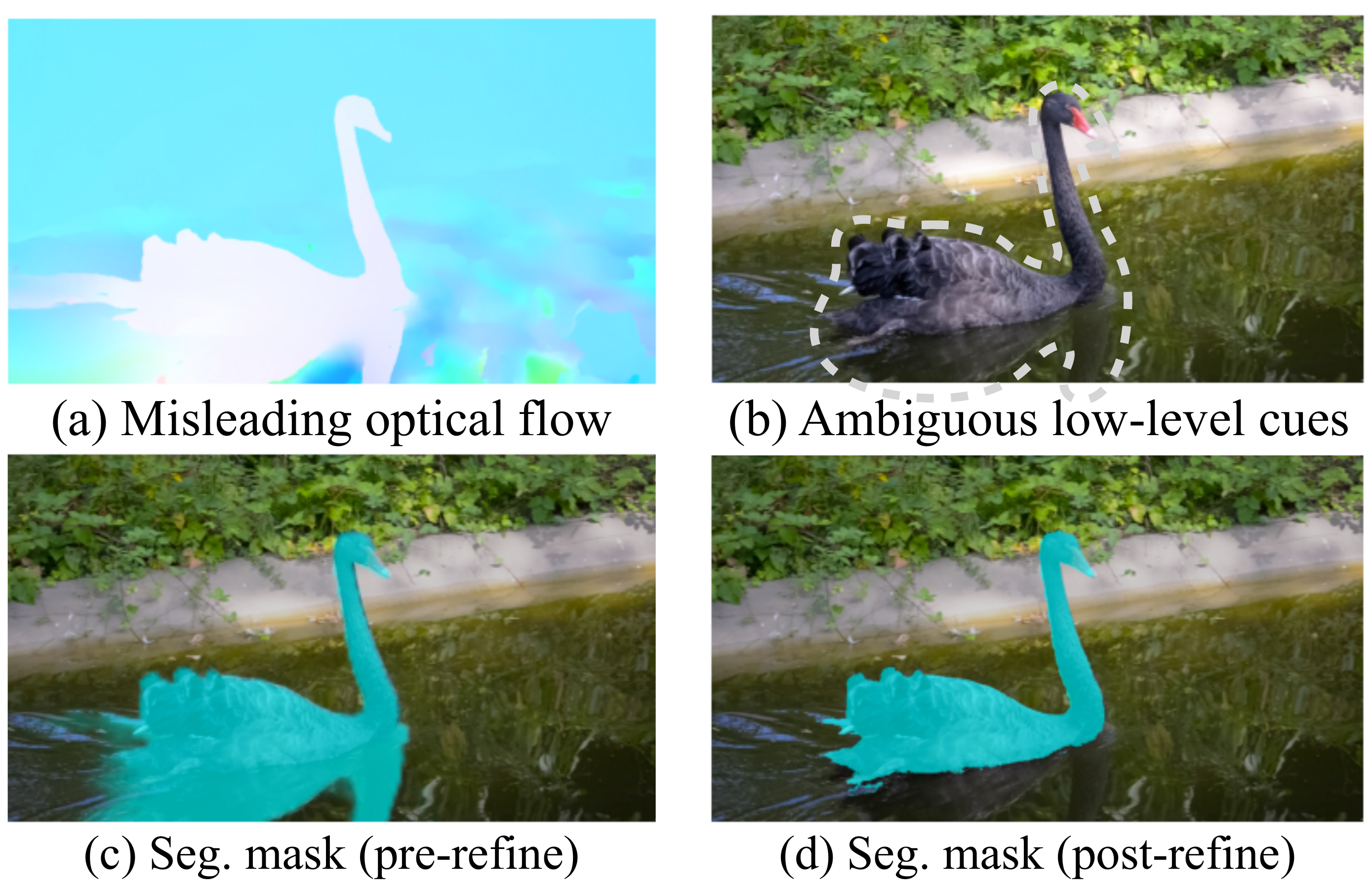}}
\vspace{-5pt}
\caption{\textbf{Semantic constraint mitigates false positives from naturally-occurring misleading motion signals.} The reflection has semantics distinct from the main object and is thus filtered out. The refined mask is then used as supervision to disperse the misconception in stage 2. Best viewed in color and zoom in. }
\vspace{-15pt}
\label{fig:refine-example}
\end{figure}
}

\def\figTuning#1{
\begin{figure}[#1]
\vspace{-5pt}
\centerline{\includegraphics[width=1.0\linewidth]{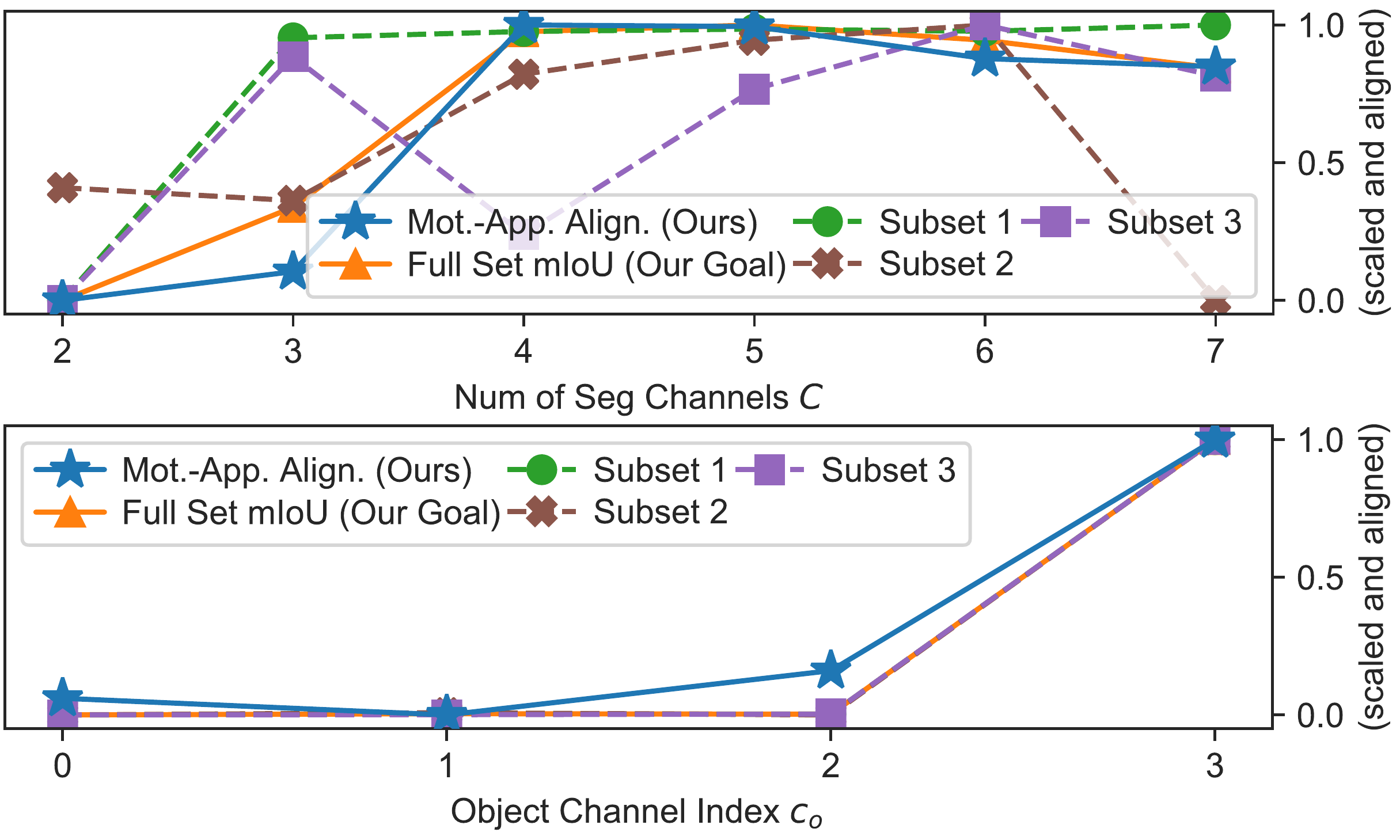}}
\vspace{-12pt}
\caption{\textbf{Our proposed label-free motion-appearance metric aligns well with mIoU on the full validation set.} \textbf{Top:} When tuning the number of segmentation channels $C$, our method follows full validation set mIoU better than mIoU on validation subsets with 25\% of the sequences labeled. \textbf{Bottom:} Our method correctly determines the object channel $c_o=3$ for this run, without any human labels. Although $c_o$ varies in each training run by design \cite{liu2021emergence}, our tuning method has negligible overhead and can be performed after training ends to find $c_o$ \textit{within seconds}.\vspace{-5pt}}
\label{fig:tuning}
\end{figure}
}

\def\figVisualizations#1{
\begin{figure*}[#1]
\vspace{-25pt}
\centering
\includegraphics[width=1.0\textwidth]{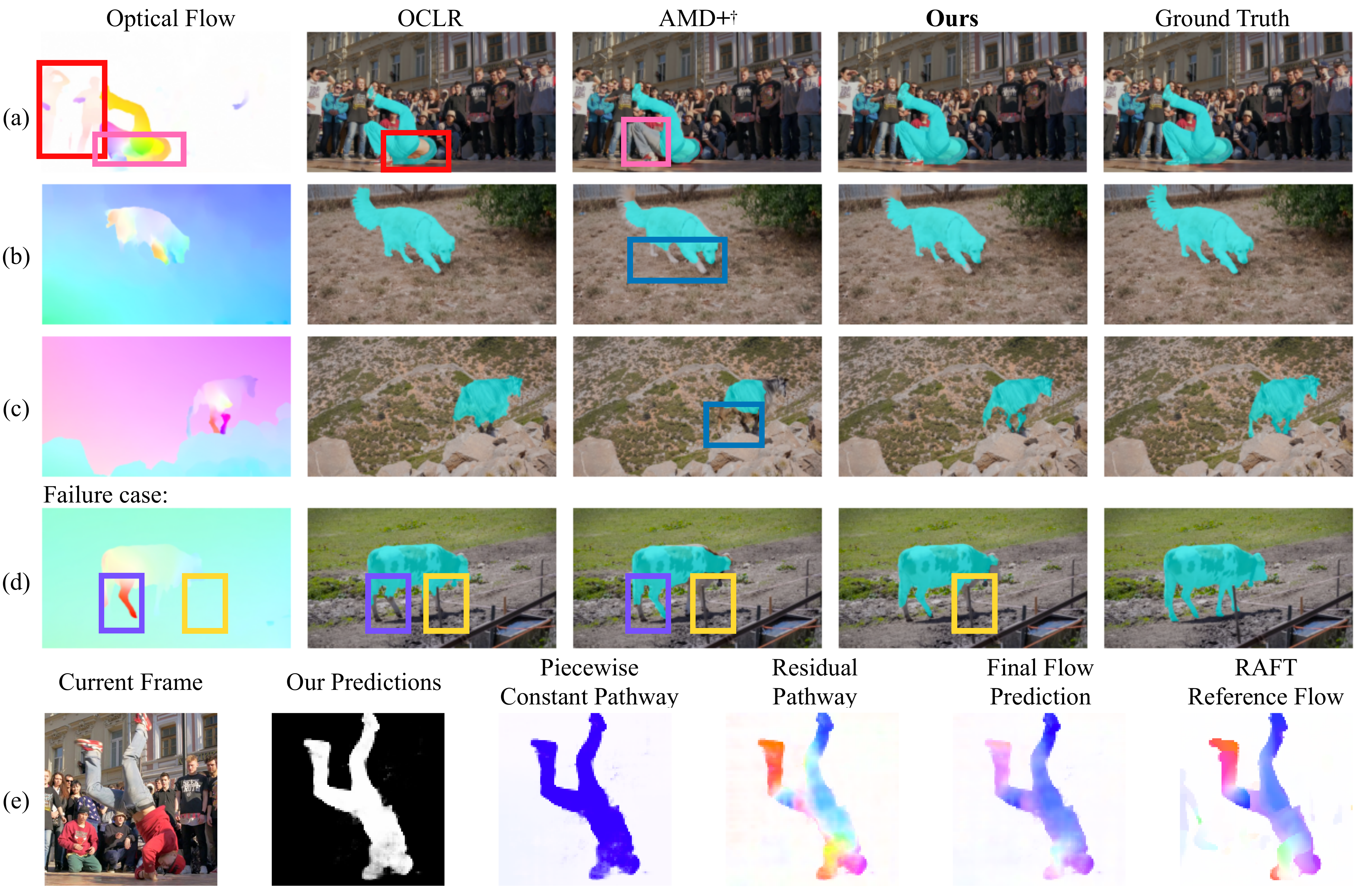}
\vspace{-15pt}
\caption{\textbf{Our method delivers great performance in challenging scenes.} Our method shows significant improvements compared to OCLR \cite{xie2022segmenting} and AMD \cite{liu2021emergence} in scenes with complex foreground motion (a)(b), distracting background motion (a)(c), motion parallax from camera motion (c). In the failure case (d), neither motion nor appearance information is informative, leading to the front legs being missed from the segmentation. However, our method still outperforms previous works and segments most of the cow's hind legs. (e) shows that the piecewise-constant pathway and the residual pathway work together to fit the reference flow, resulting in high-quality segmentation. The symbol $\dagger$ denotes AMD with RAFT flow \cite{teed2020raft} as motion supervision. More visualizations are available in the supp. mat. \vspace{-5pt}}
\label{fig:visualizations}
\end{figure*}
}

\figMotionSup{!t}

\section{Related Work}
\label{related_work}
\textbf{Unsupervised video object segmentation} (UVOS) requires segmenting prominent objects from videos  without human annotations. Mainstream benchmarks \cite{perazzi2016benchmark,li2013video,brox2010freiburg,ochs2013segmentation} define the task as a binary figure-ground segmentation problem, where salient objects are the foreground. Despite the name, several previous UVOS methods require \textit{supervised} \mbox{(pre-)training} on {\it other} data such as large-scale images or videos {\it with} manual annotations \cite{lu2019see,koh2017primary,faktor2014video,zhen2020learning,li2018unsupervised,yang2021dystab,ren2021reciprocal,zhou2020motion}. In contrast, we focus on UVOS methods which do not rely on any labels at either \textit{training or inference} time.

\textbf{Motion segmentation}  separates figure from ground based on motion, which is typically optical flow computed from a pre-trained model.
FTS \cite{papazoglou2013fast} utilizes motion boundaries for segmentation. SAGE \cite{wang2017saliency} additionally considers edges and saliency priors jointly with motion. CIS \cite{yang2019unsupervised} uses independence between foreground and background motion as the goal for foreground segmentation. However, this assumption does not always hold in real-world motion patterns. MG \cite{yang2021self} leverages attention mechanisms to group pixels with similar motion patterns. SIMO \cite{lamdouar2021segmenting} and OCLR \cite{xie2022segmenting} generate synthetic data for segmentation supervision, with the latter supporting individual segmentation of multiple objects. Nevertheless, both rely on human-annotated sprites for realistic shapes in artificial data synthesis. Motion segmentation fails when objects do not move.

\textbf{Motion-guided image segmentation} treats motion computed by a pre-trained optical flow model such as RAFT \cite{teed2020raft} as ground-truth and uses it to supervise appearance-based image segmentation.  
GWM \cite{choudhury2022guess} 
assumes smooth flows within an object and
learns appearance-based segmentation by seeking the best segment-wise affine flows that fit RAFT flows.   Such methods can discover stationary objects in videos and single images.

{\bf Joint appearance segmentation and motion estimation} methods such as AMD \cite{liu2021emergence} learn motion and segmentation simultaneously
in a self-supervised manner such that their outputs can be used to successfully reconstruct the next frame based on how segments of the current frame move.  

AMD is unique in that it has no preconception of optical flow  or visual saliency.  Since our model considers bootstrapping objectness from optical flow, for fair comparisons, we consider {AMD+}, a version of AMD with motion supervision from RAFT flows \cite{teed2020raft} instead.

Existing UVOS methods, whether they examine motion only or together with appearance,  assume that objectness is revealed through common fate of motion: What move at the same speed belong together.  
We show that this notion fails for objects with articulation and reflection (Fig.~{\ref{fig:commonFateTeaser}}).
Our RCF first bootstraps objectness by relaxed common fate and then improves it by visual appearance grouping.

\def\tabMainComparison#1{
\begin{table}[#1]
\setlength{\tabcolsep}{2.5pt}
\centering
\begin{tabular}{l||c@{}ccc}
\shline
Methods & Post-process & \textbf{DAVIS16} & \textbf{STv2} & \textbf{FBMS59} \\
\shline
SAGE \cite{wang2017saliency}          & & 42.6 & 57.6 & 61.2 \\ %
CUT \cite{keuper2015motion}           & & 55.2 & 54.3 & 57.2 \\ %
FTS \cite{papazoglou2013fast}         & & 55.8 & 47.8 & 47.7 \\ %
EM \cite{meunier2022driven}           & & 69.8 & --   & --   \\ %
CIS \cite{yang2019unsupervised}       & & 59.2 & 45.6 & 36.8 \\
MG \cite{yang2021self}                & & 68.3 & 58.6 & 53.1 \\
AMD \cite{liu2021emergence}           & & 57.8 & 57.0 & 47.5 \\
SIMO \cite{lamdouar2021segmenting}    & & 67.8 & 62.0 & -- \\
GWM \cite{choudhury2022guess}         & & 71.2 & 66.7 & 60.9 \\
\gray{GWM$^*$ \cite{choudhury2022guess}} & & \gray{71.2} & \gray{69.0} & \gray{66.9} \\
OCLR$^\dagger$ \cite{xie2022segmenting}  & & 72.1 & 67.6 & 65.4 \\
TokenCut \cite{wang2022tokencut}      & & 64.3 & 59.6 & 60.2 \\
MOD \cite{ding2022motion}             & & 73.9 & 62.2 & 61.3 \\
\hline
\textbf{RCF}      & & \textbf{80.9} & \textbf{76.7} & \textbf{69.9} \\
                   & & \textcolor{Green}{\bf\small (+7.0)} & \textcolor{Green}{\bf\small (+9.1)} & \textcolor{Green}{\bf\small (+4.5)} \\
\shline
CIS \cite{yang2019unsupervised}                & CRF + SP$^\ddagger$                & 71.5        & 62.0        & 63.6 \\
TokenCut \cite{wang2022tokencut} & CRF only & 76.7 & 61.6 & 66.6 \\
\gray{GWM$^*$} \cite{choudhury2022guess}       & \gray{CRF + SP$^\ddagger$} & \gray{73.4} & \gray{72.0} & \gray{68.6} \\
\gray{OCLR$^\dagger$} \cite{xie2022segmenting} & \gray{DINO-based$^\ddagger$} 
                                 & \gray{78.9} & \gray{71.6} & \gray{68.7} \\
\gray{MOD} \cite{ding2022motion} & \gray{DINO-based$^\ddagger$} 
                                 & \gray{79.2} & \gray{69.4} & \gray{66.9} \\
\hline
\textbf{RCF} (w/o SC) & CRF only & 82.0 & 78.7 & 71.9 \\
\textbf{RCF}      & CRF only & \textbf{83.0} & \textbf{79.6} & \textbf{72.4} \\
                   & & \textcolor{Green}{\bf\small (+6.3)} & \textcolor{Green}{\bf\small (+12.0)} & \textcolor{Green}{\bf\small (+5.8)} \\
\shline
\end{tabular}
\caption{\textbf{Our method achieves significant improvements over previous methods on common UVOS benchmarks}. RCF (w/o SC) indicates low-level refinement only (no $f_\text{aux}$ used).
$*$: uses Swin-Transformer w/ MaskFormer \cite{liu2021swin, cheng2021maskformer} segmentation head orthogonal to VOS method and thus is not a fair comparison with us.
$\dagger$ leverages manually annotated shapes from large-scale Youtube-VOS \cite{xu2018youtube} to generate synthetic data. 
$\ddagger$: \textit{SP}: significant post-processing (\eg, multi-step flow, multi-crop ensemble, and temporal smoothing). \textit{DINO-based}: performs contrastive learning or mask propagation on a pretrained DINO ViT model \cite{caron2021emerging,dosovitskiy2020image} at test time; not a fair comparison with us.
Our post-processing is a \textit{CRF pass only}. CIS results are from \cite{liu2022learning}.
\vspace{-20pt}}

\label{tab:main-comparison}

\end{table}
}

\def\tabAblation#1{
\begin{table}[#1]
\setlength{\tabcolsep}{5pt}
\centering
\begin{tabular}{ccccl}
\shline
\begin{tabular}[c]{@{}c@{}}Residual\\pathway\end{tabular} &
\begin{tabular}[c]{@{}c@{}}Low-level\\refinement\end{tabular} &
\begin{tabular}[c]{@{}c@{}}Semantic\\constraint\end{tabular}  & 
CRF & \multicolumn{1}{c}{$\mathcal{J}$ ($\uparrow$)} \\
\shline
       &         &        &        & 71.1 \\
\cmark &         &        &        & 78.9 {\small \textcolor{Green}{(+7.8)}} \\
\cmark &  \cmark &        &        & 79.9 {\small \textcolor{Green}{(+8.8)}} \\
\cmark &  \cmark & \cmark &        & 80.9 {\small \textcolor{Green}{(+9.8)}} \\
\cmark &  \cmark & \cmark & \cmark & \textbf{83.0} {\small \textbf{\textcolor{Green}{(+11.9)}}} \\
\shline
\end{tabular}
\caption{\textbf{Effect of each component of our method (DAVIS16).} Residual pathway on its own provides the most improvement in our method. %
All components together contribute to an $11.9\%$ gain.}
\label{tab:ablation}
\end{table}
}

\def\tabAblationResidual#1{
\begin{table}[#1]
\setlength{\tabcolsep}{17.5pt}
\centering
\vspace{-5pt}
\begin{tabular}{lc}
\shline
\multicolumn{1}{c}{Variants} & DAVIS16 $\mathcal{J}$ ($\uparrow$) \\
\shline
None & 71.1 \\
None (w/ robust loss \cite{sun2017pwc}) & 74.0 \\
Scaling & 73.8 \\
Residual (affine) & 76.3 \\
Residual & \textbf{78.9} \\
\shline
\end{tabular}
\vspace{-5pt}
\caption{\textbf{Ablations on additional pathway confirm our design choice of residual pathway.} We benchmark without the refinement stage to show the raw performance gain.\vspace{-5pt}}
\label{tab:ablation-residual}
\end{table}
}

\def\tabAblationCameraMotion#1{
\begin{table}[#1]
\setlength{\tabcolsep}{14.5pt}
\centering
\vspace{-5pt}
\begin{tabular}{lcc}
\shline
Camera motion modeling             & No            & Yes  \\
\shline
DAVIS16 $\mathcal{J}$ ($\uparrow$) & \textbf{78.9} & 77.9 \\
\shline
\end{tabular}
\vspace{-5pt}
\caption{\textbf{Modeling camera motion does not improve our method.} Lower segmentation quality results from removing camera motion as preprocessing. Only stage 1 is used in both settings.}
\vspace{-15pt}
\label{tab:ablation-camera-motion}
\end{table}
}

\def\tabAblationCRF#1{
\begin{table}[#1]
\setlength{\tabcolsep}{9.1pt}
\centering
\begin{tabular}{l@{\hskip -2pt}cc}
\shline
DAVIS16 $\mathcal{J}$ ($\uparrow$) & Stage 1 only & Stage 1 \& 2 \\
\shline
Without post-processing & 78.9 & 80.9 \\
With CRF post-processing & \textbf{81.4} & \textbf{83.0} \\
\hline
$\Delta$ & {\textcolor{Green}{+2.5}} & {\textcolor{Green}{+2.1}} \\
\shline
\end{tabular}
\caption{\textbf{The refinement CRF in our stage 2 is orthogonal to upsampling CRF in post-processing}, since the latter still gives significant improvements even with CRF in stage 2.\vspace{-5pt}}
\label{tab:ablation-crf}
\end{table}
}

\def\tabCompareWithSupervisedVOS#1{
\begin{table}[#1]
\setlength{\tabcolsep}{0.5pt}
\centering
\begin{tabular}{lcccccccccccc}
\shline
  & {\color{Blue}DATA}  & {\color{Blue}TMO} & {\color{Blue}RTNet} & {\color{Blue}D2Conv3D} & {\color{Blue}PMN}  & {\color{Blue}IMP} \\
\shline
$\mathcal{J}$\!\!\! & \textbf{87.1} & 85.6 & 85.6  & 85.5     & 85.4 & 84.5 \\
\shline
  & {\color{Blue}TransportNet} & {\color{Blue}AMC-Net} & {\color{Blue}3DC-Seg} & {\color{Blue}FSNet} & {\color{Blue}DFNet} & {\color{Brown}\textbf{Ours}} \\
\shline
$\mathcal{J}$\!\!\! & 84.5 & 84.5 & 84.3 & 83.4 & 83.4 & 83.0 (-0.4) \\
\shline
\end{tabular}
\caption{\textbf{RCF as an {\color{Brown}\textit{unsupervised}} VOS method achieves performance close to {\color{Blue}\textit{supervised}} VOS on DAVIS 2016.}\vspace{-5pt}}
\label{tab:compare-with-supervised-vos}
\end{table}
}

\figAppearanceSup{t!}
\section{Objectness from Relaxed Common Fate}
\label{method}

Our RCF consists of two stages: a motion-supervised object discovery stage (\cref{fig:motion-sup}) and an appearance-supervised refinement stage (\cref{fig:appearance-sup}).
Stage 1 formalizes relaxed common fate and learns segmentation by fitting RAFT flow with both object-level motion and intra-object motion.  
Stage 2 refines Stage 1's motion-based segmentations by appearance-based visual grouping and then use them to further supervise segmentation learning. 
Neither stage requires any annotation, making RCF \textit{fully unsupervised}.
We also present motion-appearance alignment as a model-agnostic label-free hyperparameter tuner.

\subsection{Problem Setting}
\label{sec:method_problem_setting}

Let $I_t \in \mathbb{R}^{3 \times h \times w}$ be the $t^{\text{th}}$ frame from a sequence of $T$ RGB frames, where $h$ and $w$ are the height and width of the image respectively. We will omit the subscript $t$ except for input images for clarity. The goal of UVOS is to produce a binary segmentation mask $M \in \{0, 1\}^{h \times w}$ for each time step $t$, with $1$ ($0$) indicating the foreground (background).

To evaluate a method on UVOS, we compute the mean Jaccard index $\mathcal{J}$ (\ie, mean IoU) between the predicted segmentation mask $M$ and the ground truth $G$. In UVOS, the ground truth mask $G$ is not available, and no human-annotations are used throughout training and inference.

\subsection{\fontsize{10.5}{10.5}\selectfont\bf\mbox{Object Discovery with Motion Supervision}}
\label{sec:method_motion_supervision}
As shown in \cref{fig:motion-sup}, during training, our method takes a pair of consecutive frames and RAFT flow between them as inputs. To instantiate the idea of common fate, our method begins by pooling the RAFT Flow with respect to the predicted masks, creating the piecewise constant flow pathway. As a relaxation, the predicted residual flow, which models intra-object motion for articulated and deformable objects, is added to the piecewise constant flow. The composite flow prediction is then supervised by the RAFT flow to train the model. At test time, only the backbone and the segmentation head are utilized to perform inference per frame.

Specifically, let $f(I_t) \in \mathbb{R}^{K \times H \times W}$ be the feature of $I_t$ extracted from backbone $f(\cdot)$, where $K$, $H$, and $W$ are the number of channels, height, and width of the feature. Let $\hat M = g(f(I_t)) \in \mathbb{R}^{C \times H \times W}$ be $C$ soft segmentation masks extracted with a lightweight fully convolutional segmentation head $g(\cdot)$ taking the image feature from $f(\cdot)$. \Verb/Softmax/ is taken across channels inside $g(\cdot)$ so that the $C$ soft masks sum up to 1 for each of the $H \times W$ positions. Following \cite{liu2021emergence}, although there are $C$ segmentation masks competing for each pixel (i.e., $C$ output channels in $\hat M$), \textit{only one} corresponds to the foreground, with the rest capturing background patches. We define $c_o$ as the object channel index with value obtained in \cref{sec:method_tuning}.

Following \cite{yang2021self,lamdouar2021segmenting,xie2022segmenting,choudhury2022guess}, we use off-the-shelf optical flow model RAFT \cite{teed2020raft} trained on synthetic datasets\cite{dosovitskiy2015flownet,mayer2016large} to provide motion cues between consecutive frames. Let $F \in \mathbb{R}^{2 \times H \times W}$ be the flow output from RAFT from $I_t$ to $I_{t+1}$.

\noindent\textbf{Piecewise constant pathway.} We first pool the flow according to each mask to form $C$ flow vectors $\hat P_{c} \in \mathbb{R}^{2}$:
\vspace{-2pt}
\begin{align}
&\hat P_{c} = \phi_2(\text{GuidedPool}(\phi_1(F), \hat M_{c}))\\
&\text{GuidedPool}(F, M) = \frac{\sum_{p=1}^{HW} (F \odot M)[p]}{\sum_{p=1}^{HW}M[p]}
\end{align}
where $[p]$ denotes the pixel index and $\odot$ element-wise multiplication. Following \cite{liu2021emergence}, $\phi_1$ and $\phi_2$ are two-layer lightweight MLPs that transform each of the motion vectors independently before and after pooling, respectively.
We then construct predicted flow $\hat P \in \mathbb{R}^{2 \times H \times W}$ according to the soft segmentation mask:
\vspace{-2pt}
\begin{align}
&\hat P = \sum_{c=1}^{C} \text{Broadcast}(\hat P_{c}, \hat M_{c}) \\
&\text{Broadcast}(\hat P_{c}, \hat M_{c})[p] = \hat P_{c} \odot (\hat M_{c}[p]).
\end{align}
As the mask prediction $\hat M_c$ approaches binary during training, the flow prediction approaches a piecewise-constant function with respect to each segmentation mask, capturing common fate. Previous methods either directly supervise $\hat P$ with an image warping for self-supervised learning \cite{liu2021emergence} or matches $\hat P$ and $F$ by minimizing the discrepancies up to an affine factor (\ie, up to first order) \cite{choudhury2022guess}.

Nonetheless, hand-crafted non-learnable motion models, while capturing the notion of common fate, underfit complex optical flow in real-world videos, which often put object parts into different segmentation channels in order to minimize the loss, despite similar color or texture. \cite{choudhury2022guess} uses two mask channels as a remedy, still falling short for scenes with complex backgrounds.

\noindent\textbf{Learnable residual pathway.} 
Rather than using more complicated hand-crafted motion models to model the motion patterns in videos, we employ \textit{relaxed} common fate by separately fitting object-level and intra-object motion by adding a \textit{learnable} residual pathway $\hat R$ in addition to the piecewise constant pathway $\hat P$ to form the final flow prediction $\hat F$. The residual pathway models relative intra-object motion such as the relative motion of the dancer's feet to the body in \cref{fig:motion-sup}.

Let $h(\cdot)$ be a lightweight module with three Conv-BN-ReLU blocks that take the concatenated feature of a pair of frames $\{I_t, I_{t+1}\}$ as input and predicts $\hat R' \in \mathbb{R}^{C \times 2 \times H \times W}$, which includes $C$ flows with per-pixel upper bound $\lambda$:
\vspace{-2pt}
\begin{align}
\hat R' = \lambda\tanh(h(\text{concat}(f(I_t), f(I_{t+1})))
\end{align}
where the upper bound $\lambda$ is set to $10$ pixels unless stated otherwise.
The $C$ residual flows form aggregated residual flow $\hat R$ using mask predictions, which sums up with the piecewise constant pathway to form the final flow prediction $\hat F$:
\vspace{-10pt}
\begin{align}
\hat R &= \sum_{c=1}^{C} \hat R_{c}' \odot \hat M_{c} \\
\hat F &= \hat P + \hat R
\end{align}
In this way, $\hat F$ additionally takes into account relative motion that is within $(-\lambda, \lambda)$ for each spatial location. 
The added residual pathway provides greater flexibility by allowing the model to relax from common fate that does not take intra-object motion into account. This leads to better segmentation results for articulated and deformable objects.

At stage 1, we minimize the L1 loss between the predicted reconstruction flow $\hat F$ and target flow $F$ in order to learn segmentation by predicting the correct flow:
\vspace{-5pt}
\begin{equation}
L_{\text{stage 1}} = L_{\text{motion}} = \frac{1}{HW}\sum_{p=1}^{HW}||\hat F[p] - F[p]||_1
\end{equation}
\vspace{-10pt}

\subsection{\mbox{Refinement with Appearance Supervision}}
\label{sec:method_refinement}
A primary focus of self-supervised learning is to find sources of useful training signals. While the residual pathway greatly improves segmentation quality, the supervision still primarily comes from motion. This single source of supervision can lead to predictions that are optimal for flow prediction but often suboptimal from an appearance perspective. For instance, in \cref{fig:appearance-sup}, the segmentation prediction before refinement ignores a part of the dancer's leg, despite the ignored part sharing a very similar color and texture with the included parts. Furthermore, the RAFT flow tends to be noisy in areas where nearby pixels move very differently, which leads to segmentation ambiguity.

To address these issues, we propose to incorporate low- and high-level appearance signals as another source of supervision to correct the misconceptions from motion.

\noindent\textbf{Appearance supervision with low-level intra-image cues.} With the model in stage 1, we obtain the mask prediction $\hat M_{c_o}$ of $I_t$, where $c_o$ is the objectness channel that could be found without annotation (\cref{sec:method_tuning}). We then apply fully-connected conditional random field~(CRF) \cite{krahenbuhl2011efficient}, a training-free technique that refines the value of each prediction based on other pixels with an appearance and a smoothness kernel.
The refined masks $\hat M'_{c_o}$ are then used as supervision to provide appearance signals in training:
\vspace{-5pt}
\begin{align}
\hat M'_{c_o} &= \text{CRF}(\hat M_{c_o}) \label{eq:crf_refinement}\\[-2pt]
L_{\text{app}} &= \frac{1}{HW} \sum_{p=1}^{HW}||\hat M_{c_o}[p] - \hat M'_{c_o}[p]||_2^2
\end{align}
\vspace{-12pt}

The total loss in stage 2 is a weighted sum of both motion and appearance loss:
\vspace{-5pt}
\begin{equation}
L_\text{stage 2} = w_\text{app} L_\text{app} + w_\text{motion} L_\text{motion} 
\end{equation}
\vspace{-18pt}

\noindent where $w_\text{app}$ and $w_\text{motion}$ are weights for loss terms.

The CRF in our method for appearance supervision is different with the traditional CRF used in post-processing~\cite{yang2019unsupervised,choudhury2022guess}, as our refined masks provide the supervision for training the network. Furthermore, we show empirically that our method is orthogonal to the traditional CRF in the ablation (\cref{sec:ablation}).

\figRefineExample{!t}
\noindent\textbf{Appearance supervision with semantic constraint.} %
Low-level appearance is still insufficient to address misleading motion signals from naturally occurring confounders with similar motion patterns. 
For example, the reflections share similar motion as the swan in \cref{fig:refine-example}, which is confirmed by low-level appearance. 
However, humans could recognize that the swan and the reflection have distinct semantics, with the reflection's semantics much closer to the background. %

Inspired by this, we incorporate the statistically learned feature map from a frozen auxiliary DINO ViT\cite{caron2021emerging,dosovitskiy2020image} trained with self-supervised learning across ImageNet \cite{ILSVRC15} without human annotation, to create a semantic constraint for mask prediction. %
We begin by taking the key features from the last transformer layer, denoted as $f_\text{aux}(I_t)$, inspired by \cite{wang2022tokencut}. Next, we compute and iteratively optimize the normalized cut \cite{shi2000normalized} with respect to the mask to refine the mask.

Specifically, we initialize a 1-D vector $\vec{x}$ with a flattened and resized $\hat M_{c_o}$ with shape $HW$. Then we build an appearance-based affinity matrix $A$, where:
\vspace{-2pt}
\begin{align}
A_{ij}=\mathbb{1}(\text{sim}(f_\text{aux}(I_t)_i, f_\text{aux}(I_t)_j) \geq 0.2)
\end{align}

Next, we compute $\text{NCut}(A, \vec{x})$:
\vspace{-2pt}
\begin{align}
\text{Cut}(A, \vec{x}) &= (1-\vec{x}) A \vec{x} \\
\text{NCut}(A, \vec{x}) &= \frac{\text{Cut}(A, \vec{x})}{\sum_{i=1}^{HW} (A\vec{x})_i} 
+ \frac{\text{Cut}(A, \vec{x})}{\sum_{i=1}^{HW} (A(1-\vec{x}))_i} 
\end{align}
where $\text{sim}(\cdot, \cdot)$ cosine similarity. Since $\text{NCut}(A, \vec{x})$ is differentiable with respect to $\vec{x}$, we use Adam\cite{kingma2014adam} to minimize $\text{NCut}(A, \vec{x})$ in order to refine $\vec{x}$ for $k=10$ iterations. We denote the optimized vector as $\vec{x}^{(k)}$, which is thus the refined version of the mask that carries consistent semantics, thus decoupling the objects from their shadows and reflections. With the semantic constraint, \cref{eq:crf_refinement} changes to:
\vspace{-2pt}
\begin{equation}
\hat M'_{c_o} = \text{CRF}(\hat M_{c_o}) \odot \text{CRF}(\vec{x}^{(k)})
\end{equation}
where $\vec{x}^{(k)}$ is reshaped to 2D and resized to match the mask sizes prior to CRF. Since stage 2 is mainly misconception correction and thus much shorter than stage 1, we generate the NCut refined masks only once and use the same refined masks for efficient supervision.

Because the semantic constraint introduces an additional frozen model $f_\text{aux}(\cdot)$, we benchmark both \textit{with} and \textit{without} the semantic constraint for a fair comparison with previous methods. We use \textit{\textbf{RCF} (w/o SC)} to denote RCF without the semantic constraint. Our method is still fully unsupervised even with the semantic constraint.

\subsection{Label-free Hyperparameter Tuner}
\label{sec:method_tuning}
Following previous appearance-based UVOS work, our method also requires several tunable hyperparameters for high-quality segmentation. The most critical ones are the number of segmentation channels $C$ and the object channel index $c_o$.
\cite{liu2021emergence,choudhury2022guess} tune both hyperparameters either with a large labeled validation set or a hand-crafted metric tailored to a specific hyperparameter, limiting the capability towards other hyperparameters in a real-world setting.

We propose motion-appearance alignment as a metric to quantify the segmentation quality. The steps for tuning are:
\ol{enumerate}{-1}{
\item Train a model with each hyperparameter setting. 
\item Export the predicted mask $\hat M_{c_o}$ for each image in the \textit{unlabeled} validation set. 
\item Compute the negative normalized cuts $-\text{NCut}(A, \hat M_{c_o})$ w.r.t. $\hat M_{c_o}$ and the appearance-based affinity matrix $A$ as the metric quantifying motion-appearance alignment.
\item Take the mean metric across all validation images.
\item Select the setting with the highest mean metric.
}

Our hyperparameter tuning method is model-agnostic and applicable to other UVOS methods. We also demonstrate its effectiveness in tuning weight decay and present the pseudo-code in the supp. mat.

\section{Experiments}
\label{experiments}
\tabMainComparison{!t}
\subsection{Datasets}
We evaluate our methods using three datasets commonly used to benchmark UVOS, following previous works \cite{yang2019unsupervised,yang2021self,liu2021emergence,choudhury2022guess,xie2022segmenting}. \textbf{DAVIS2016} \cite{perazzi2016benchmark} contains 50 video sequences with 3,455 frames in total. Performance is evaluated on a validation set that includes 20 videos with annotations at 480p resolution. \textbf{SegTrackv2} (STv2) \cite{li2013video} contains 14 videos of different resolutions, with 976 annotated frames and lower image quality than \cite{perazzi2016benchmark}. \textbf{FBMS59} \cite{ochs2013segmentation} contains 59 videos with a total of 13,860 frames, 720 frames of which are annotated with a roughly fixed interval. We follow previous work to merge multiple foreground objects in STv2 and FBMS59 into one mask and train on all unlabeled videos. We adopt mean Jaccard index~$\mathcal{J}$ (mIoU) as the primary evaluation metric.

\subsection{Unsupervised Video Object Segmentation}
\noindent\textbf{Setup.} Our architecture is simple and straightforward. We use a ResNet50 \cite{he2016deep} backbone followed by a segmentation head and a residual prediction head. Both heads only consist of three Conv-BN-ReLU layers with 256 hidden units. This standard design allows efficient implementation in real-world applications. Unless otherwise stated, we use $C=4$ object channels, which we determine without human annotation in \cref{sec:experiment_tuning}. We also determine the object channel index $c_o$ using the same approach. The RAFT \cite{teed2020raft} model we use is only trained on synthetic FlyingChairs \cite{dosovitskiy2015flownet} and FlyingThings \cite{mayer2016large} dataset without human annotation. For more details, please refer to supplementary materials.

\noindent\textbf{Results.} As shown in \cref{tab:main-comparison}, RCF outperforms previous methods under fair comparison, often by a large margin. On DAVIS16, RCF surpasses the previous state-of-the-art method by $7.0\%$ without post-processing (abbreviated as pp.). With CRF as the only pp., RCF improves on previous methods by $6.3\%$ without techniques such as multi-step flow, multi-crop ensemble, and temporal smoothing. RCF also outperforms GWM\cite{choudhury2022guess} that employs more complex Swin-T + MaskFormer architecture \cite{liu2021swin,cheng2021maskformer} by $9.7\%$ w/o pp. Furthermore, RCF achieves significantly better results compared with TokenCut\cite{wang2022tokencut} that also uses normalized cuts on DINO features\cite{caron2021emerging} ($16.6\%$ better w/o pp.). Despite the varying image quality in STv2 and FBMS59, RCF improves over past methods under fair comparison, by $9.1\%$ and $4.5\%$ without pp, respectively. Semantic constraint (SC) could be included if additional gains are desired. However, RCF still outperforms previous works without the semantic constraint ($5.3\%$ improvement on DAVIS16 w/o SC), thus \textit{not relying on external frozen features}.

\subsection{Label-free Hyperparameter Tuning}
\label{sec:experiment_tuning}
We use motion-appearance alignment as a metric to tune two key hyperparameters: the number of segmentation masks $C$ and the object channel index $c_o$. To simulate the real-world scenario that we only have limited labeled validation data, we also randomly sample 25\% sequences three times to create three labeled subsets of the validation set to evaluate mIoU on. %
As shown in \cref{fig:tuning}, for \textbf{the number of mask channels $C$}, despite \textit{not using any manual annotation}, our label-free motion-appearance alignment closely follows the validation mIoU compared to mIoU on validation subsets, showing the effectiveness of our metric on hyperparameter tuning. Although increasing the number of channels improves the segmentation quality of our model by increasing its fitting power, such an increase saturates at $C=4$. Therefore, we use $C=4$ unless otherwise stated.
Regarding \textbf{the object channel index $c_o$}, since it changes with each random initialization \cite{liu2021emergence}, optimal $c_o$ needs to be obtained at the end of each training run. We propose to use only the first frame of each video sequence for finding $c_o$. With this adjustment, our tuning method completes within \textit{only 3 seconds} for each candidate channel, which enables our tuning method to be performed after the whole training run with negligible overhead.

\figTuning{!t}
\figVisualizations{!t}

\subsection{Ablation Study}
\label{sec:ablation}

\tabAblation{!t}
\tabAblationResidual{!t}
\tabAblationCRF{!t}

\noindent\textbf{Contributions of each component.} As shown in \cref{tab:ablation}, residual pathway allows more flexibility and contributes $7.8\%$ mIoU. The appearance refinement in the second stage boosts the performance to $80.9\%$, resulting in a $9.8\%$ gain in total. The CRF post-processing leads to $83.0\%$ mIoU, an $11.9\%$ increase over the baseline.

\noindent\textbf{Designing additional pathway.} In \cref{tab:ablation-residual}, we show that robustness loss \cite{sun2017pwc,liu2020learning} does not effectively reduce the impact of misleading motion. We also implemented a pixel-wise scaling pathway, which multiplies each value of the motion vector by a predicted value. Furthermore, we fit an affine transformation per segmentation channel as the residual. In our setting, the pixel-wise residual performs the best and is selected for our model, showing the effectiveness of a \textit{learnable} and \textit{flexible} motion model inspired by relative motion.

\noindent\textbf{Orthogonality of our appearance supervision with post-processing}. The supervision from refined masks after appearance-based refinement has the same resolution as the original exported masks. Therefore, the refinement CRF in stage 2 has an orthogonal effect to the upsampling CRF in post-processing mainly used to create high-resolution masks. As shown in \cref{tab:ablation-crf}, the gains that come from post-processing remain comparable after applying appearance-based refinement in stage 2, which also shows our orthogonality to post-processing.

\tabAblationCameraMotion{t!}
\noindent\textbf{Modeling camera motion?}
RCF does not explicitly model the flow from camera motion.
To investigate whether modeling camera motion could further benefit RCF, we estimate it with the planar homography and RANSAC \cite{fischler1981random} and remove it as a preprocessing step prior to training our method. Despite the relatively accurate estimation when visualized, \cref{tab:ablation-camera-motion} shows that it is ineffective in improving the segmentation quality. We hypothesize that it is because 3D camera motion is equivalent to 3D scene motion in an opposite direction and thus additional modeling is unnecessary.

\subsection{Visualizations and Discussions}
\cref{fig:visualizations} compares RCF with \cite{xie2022segmenting,liu2021emergence} and shows its ability to handle challenging cases such as complex non-uniform foreground motion, distracting background motion, and camera motion including rotation. However, when neither motion nor appearance provides informative signals, RCF may suffer from the lack of information. For instance, in the absence of relative motion, RCF is misled by the similarity between the color of the cow's front legs and the color of the ground in \cref{fig:visualizations}(d). Although RCF has the ability to recognize multiple foreground objects with similar motion, it sometimes captures only one object when the objects move in very different patterns. Finally, RCF is not designed to separate multiple foreground objects. More visualizations and discussions are available in the supp.~mat.

\section{Summary}
\label{summary}
We present RCF, an unsupervised video object segmentation method based on relaxed common fate and appearance grouping. Our approach includes a motion-supervised object discovery stage with a learnable residual pathway, a refinement stage with appearance supervision, and using motion-appearance alignment as a label-free hyperparameter tuning method. Extensive experiments show our method's effectiveness and utility in challenging scenarios.

\noindent\textbf{Acknowledgements.} The authors would like to thank Zilin Wang for proofreading this paper.

\def\figVisualizationsSuppResidual#1{
\begin{figure*}[#1]
\centering
\includegraphics[width=1.0\textwidth]{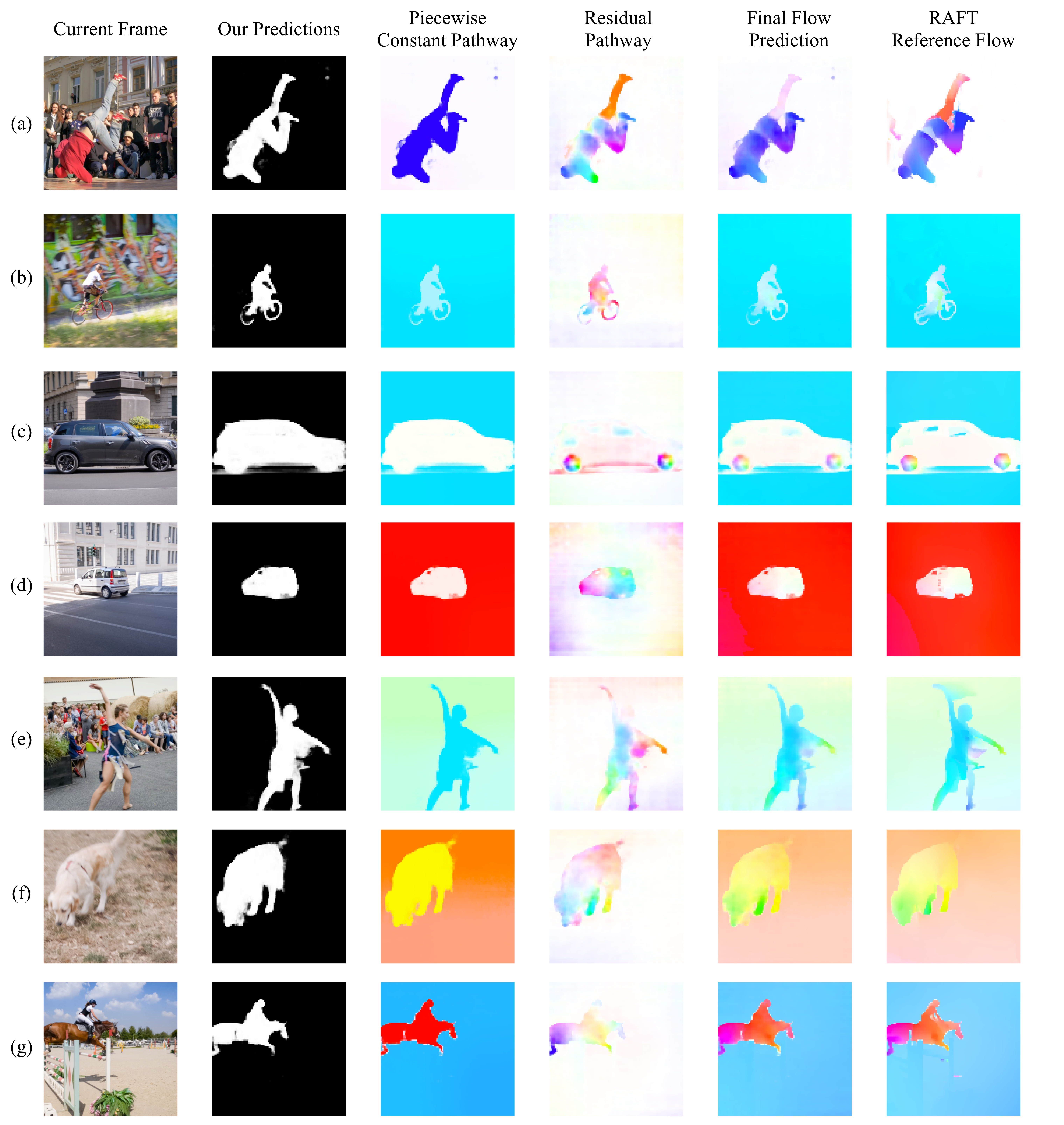}
\caption{\textbf{Visualizations for both the piecewise constant and the residual pathways} show that the introduction of the residual pathway allows our segmentation prediction to better fit the flow of deformable and articulated objects. In addition, it also relieves our segmentation module from strictly fitting the flow from 3D rotation and changing depth in a piecewise constant manner. By modeling relative motion in 2D flow, the residual pathway makes our method flexible and robust to objects with complex motion.}
\label{fig:visualizations_residual}
\end{figure*}
}

\def\figVisualizationsSuppDAVIS#1{
\begin{figure*}[#1]
\centering
\includegraphics[width=1.0\textwidth]{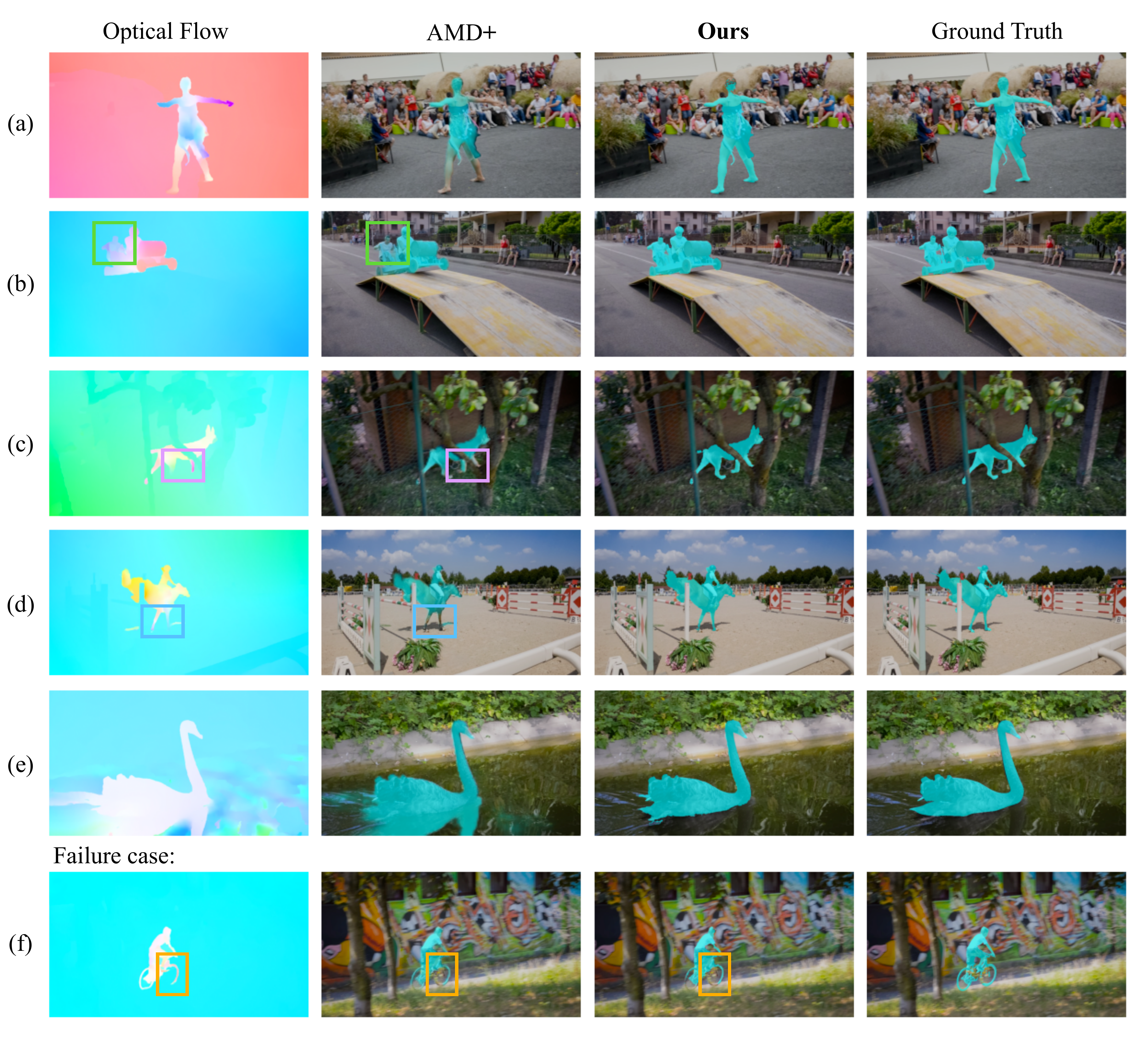}
\caption{\textbf{Additional visualizations on DAVIS16 \cite{perazzi2016benchmark}.} Our method remains robust in scenes where there is insufficient motion information, in which cases our method leverages appearance cues to learn high-quality segmentation in (a) to (e). Our method accurately segments multiple foreground objects as foreground when they move together, which is consistent with human perception in (b). However, our method may exclude a portion of an object in (f), since the motion misses part of the front wheel of the bicycle and the structure is too small for appearance to capture.}
\label{fig:visualizations_davis}
\end{figure*}
}

\def\figVisualizationsSuppST#1{
\begin{figure*}[#1]
\centering
\includegraphics[width=1.0\textwidth]{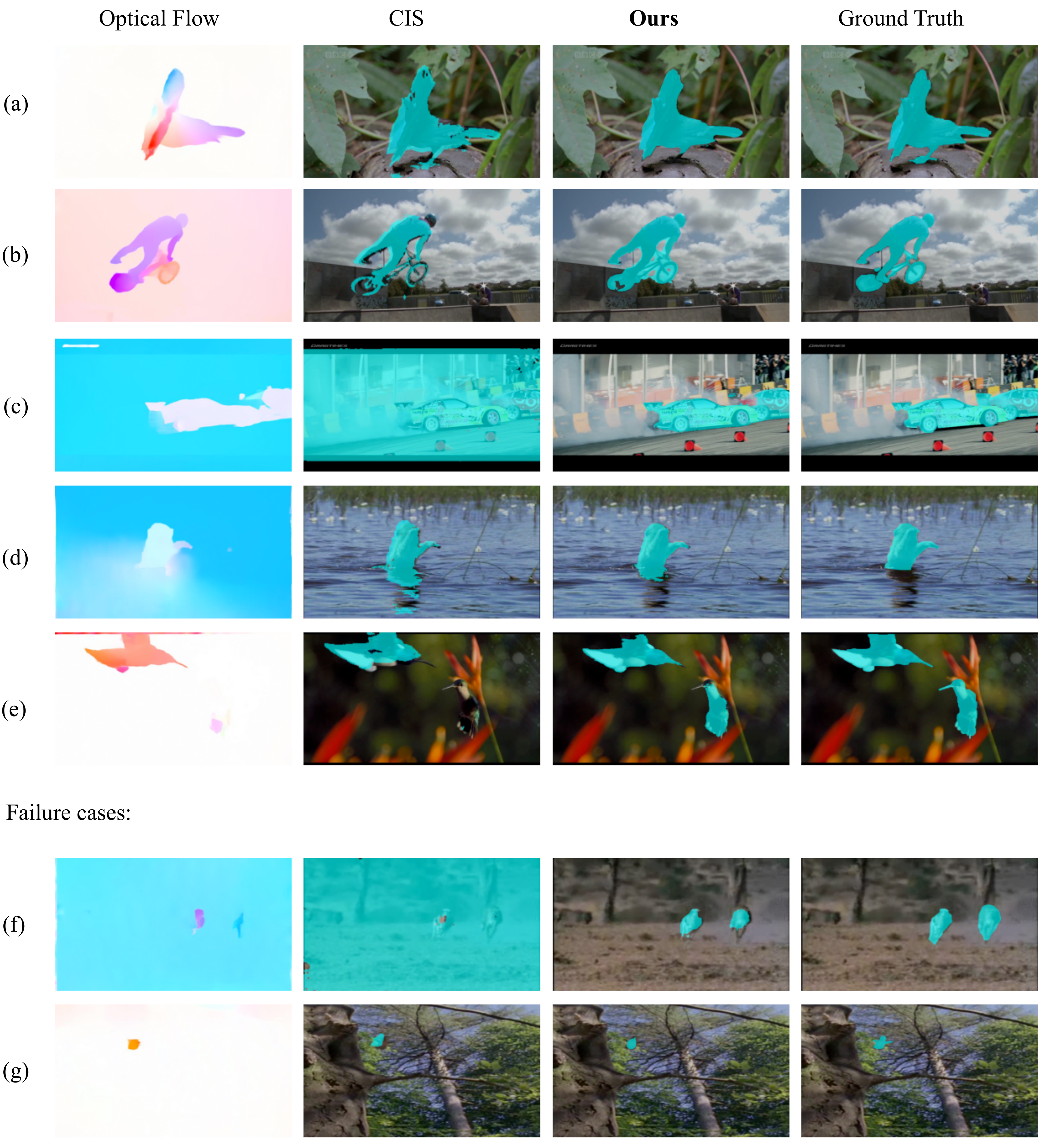}
\caption{\textbf{Additional visualizations on STv2 \cite{li2013video}.} Our method, with the residual flow, could model non-uniform 2D flow resulting from object rotation in 3D in (a), as long as the rotation flow falls within our upper bound constraint for the residual flow. Our method also captures multiple objects in a foreground group in (b), (c), and (e). Our method is robust to camera motion that leads to non-uniform background flow in (c) and misleading common motion (reflections) in (d). However, due to the relatively low image resolution, our method may miss some details of the object. For example, the legs of both animals in (f) and the wings of the bird in (g).}
\label{fig:visualizations_st}
\end{figure*}
}

\def\figVisualizationsSuppFBMS#1{
\begin{figure*}[#1]
\centering
\includegraphics[width=1.0\textwidth]{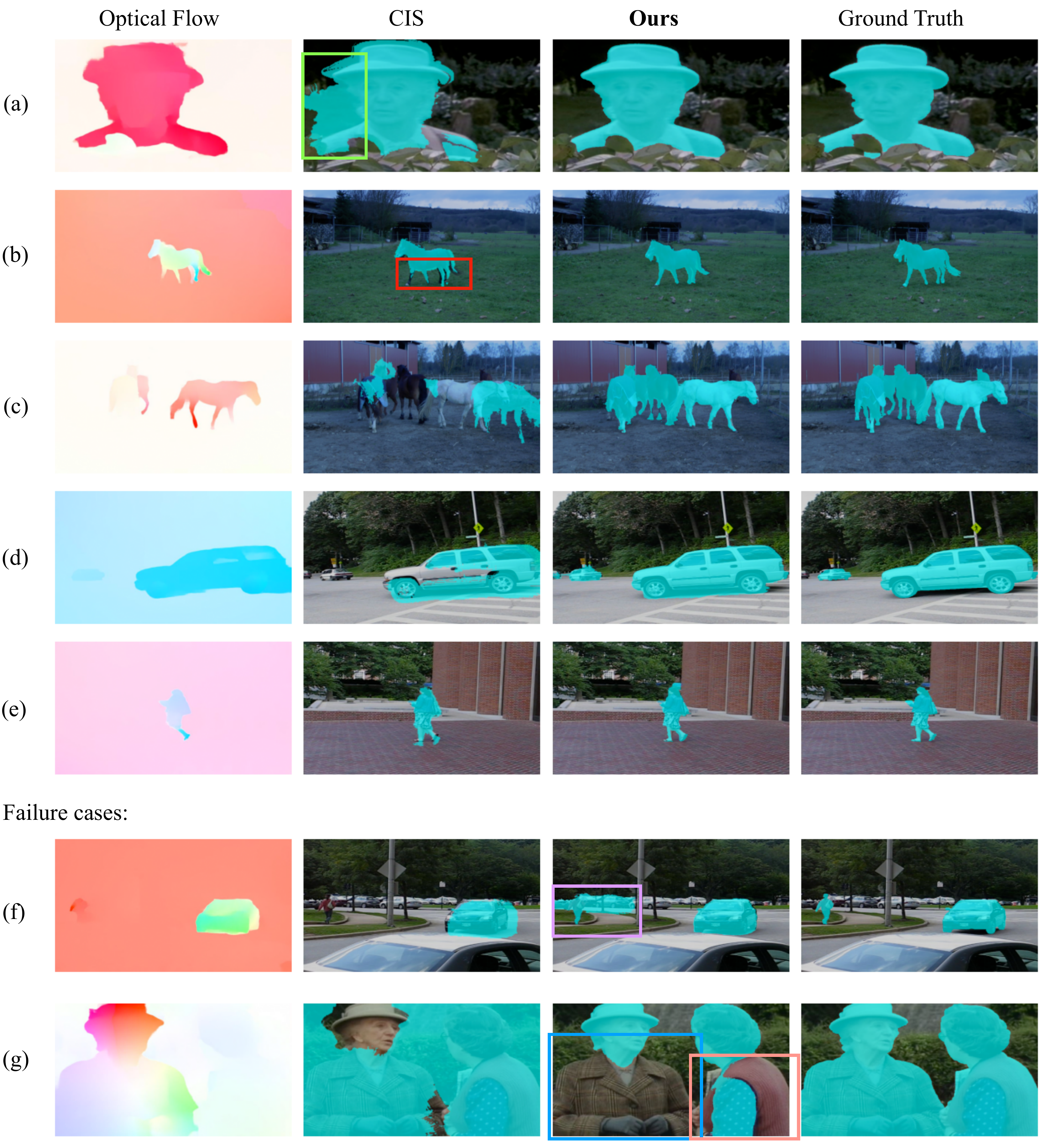}
\caption{\textbf{Additional visualizations on FBMS59 \cite{brox2010freiburg,ochs2013segmentation}.} Our method is robust in scenes with complicated and distracting appearances in (a). Our method also works with fine details in (b) and (e). Our method accurately segments multiple foreground objects in (c) and (d). However, when multiple objects or object parts exist in one scene and exhibit different motion patterns, our method may be confused in (f) and (g).}
\label{fig:visualizations_fbms59}
\end{figure*}
}

\def\figTuningAMD#1{
\begin{figure}[#1]
\centerline{\includegraphics[width=1.0\linewidth]{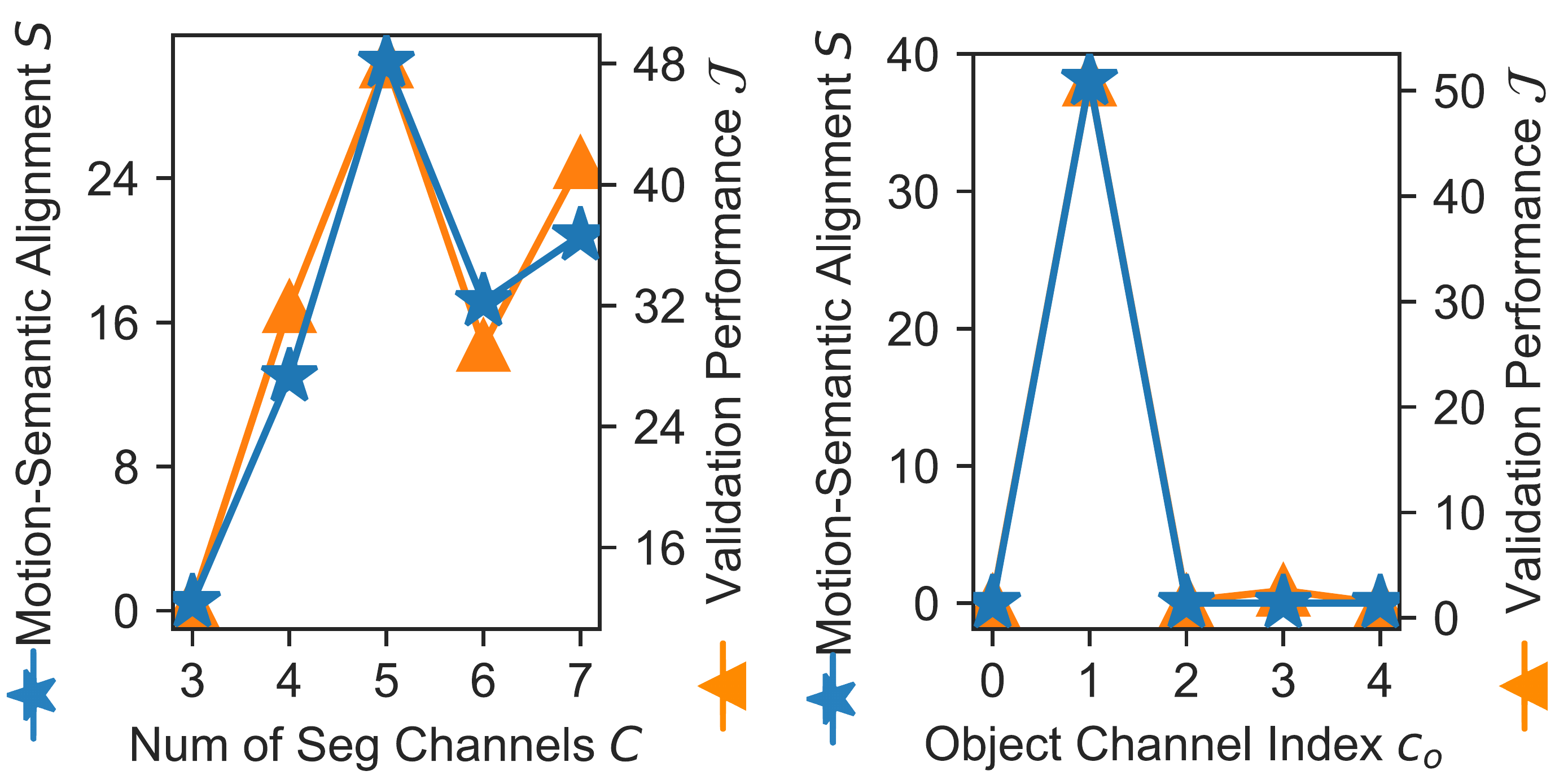}}
\caption{\textbf{Our hyperparameter tuning technique is model agnostic.} When using our model-agnostic hyperparameter tuning technique on AMD \cite{liu2021emergence}, the tuned hyperparameters from unsupervised motion-semantic alignment greatly resemble the ones obtained with human annotation. In this training run, the number of segmentation channels is 4 to be optimal, and the object channel index is 1 from \textit{both} our motion-semantic alignment and validation performance. Although $c_o$ varies in each training run by design \cite{liu2021emergence}, our tuning method has negligible overhead and could be performed after training to find $c_o$ \textit{within seconds}.
}
\label{fig:tuning_amd}
\end{figure}
}

\def\tabFlowMethod#1{
\begin{table}[#1]
\tablestyle{1.0pt}{1.0}
\centering
\begin{tabular}{c|c|c|c|c|c|c|c|c}
\shline
 Method & CIS & MG & EM & SIMO & Tok.Cut & GWM & OCLR & \textbf{RCF} \\
 \shline
Flow Model & PWCNet & RAFT & RAFT & RAFT & RAFT & RAFT & RAFT & \textbf{RAFT} \\
\shline
\end{tabular}
\caption{\textbf{Optical flow methods that each UVOS approach employs by default.} All methods in the table use pretrained weights for flow estimation. We utilize RAFT flow with pretrained weights from synthetic data, which is common among all the UVOS methods. Other than the methods listed in the table, AMD trains PWCNet \cite{sun2017pwc} architecture from scratch but achieves much lower performance compared to RCF.}
\label{tab:flow-method}
\end{table}
}

\def\tabAblationFlowMethod#1{
\begin{table}[#1]
\tablestyle{2pt}{1.0}
\centering
\begin{tabular}{c|c|c|c|c}
\shline
 Method & ARFlow \cite{liu2020learning} & PWCNet \cite{sun2017pwc} & GMFlow \cite{xu2022gmflow} & \textbf{RAFT} \cite{teed2020raft} \\
 \shline
DAVIS16 $\mathcal{J} (\uparrow)$ & 70.3 & 74.8 & 76.6 & \textbf{78.9} \\
\shline
\end{tabular}
\caption{\textbf{Our method with different optical flow estimation methods.} We use pretrained optical flow on synthetic data for supervised optical flow methods. We benchmark stage 1 only since we leverage motion supervision mostly in stage 1.}
\label{tab:ablation-flow-method}
\end{table}
}

\def\tabDAVISPerSequence#1{
\begin{table}[#1]
\setlength{\tabcolsep}{20pt}
\centering
\begin{tabular}{lc}
\shline
Sequence & $\mathcal{J}$ \\
\shline
blackswan & 76.2 \\
bmx-trees & 78.3 \\
breakdance & 86.1 \\
camel & 92.7 \\
car-roundabout & 80.7 \\
car-shadow & 80.4 \\
cows & 88.0 \\
dance-twirl & 90.4 \\
dog & 91.7 \\
drift-chicane & 94.1 \\
drift-straight & 65.6 \\
goat & 81.6 \\
horsejump-high & 93.4 \\
kite-surf & 53.1 \\
libby & 96.6 \\
motocross-jump & 57.0 \\
paragliding-launch & 26.0 \\
parkour & 95.8 \\
scooter-black & 72.4 \\
soapbox & 86.1 \\
\hline
Frame Avg & 83.0 \\
\shline
\end{tabular}
\caption{\textbf{Per sequence Jaccard index $\mathcal{J}$ on DAVIS16 \cite{perazzi2016benchmark}.}}
\label{tab:davis16_per_seq}
\end{table}
}

\def\tabSTPerSequence#1{
\begin{table}[#1]
\setlength{\tabcolsep}{20pt}
\centering
\begin{tabular}{lc}
\shline
Sequence & $\mathcal{J}$ \\
\shline
bird of paradise & 91.7 \\
birdfall & 60.4 \\
bmx & 76.6 \\
cheetah & 52.4 \\
drift & 86.3 \\
frog & 82.2 \\
girl & 80.6 \\
hummingbird & 67.6 \\
monkey & 82.5 \\
monkeydog & 55.5 \\
parachute & 93.2 \\
penguin & 66.2 \\
soldier & 79.8 \\
worm & 85.6 \\
\hline
Frame Avg & 79.6 \\
\shline
\end{tabular}
\caption{\textbf{Per sequence Jaccard index $\mathcal{J}$ on STv2 \cite{li2013video}.}}
\label{tab:stv2_per_seq}
\end{table}
}

\def\tabFBMSPerSequence#1{
\begin{table}[#1]
\setlength{\tabcolsep}{20pt}
\centering
\begin{tabular}{lc}
\shline
Sequence & $\mathcal{J}$ \\
\shline
camel01 & 88.3 \\
cars1 & 86.4 \\
cars10 & 38.2 \\
cars4 & 70.3 \\
cars5 & 79.3 \\
cats01 & 88.2 \\
cats03 & 82.0 \\
cats06 & 59.7 \\
dogs01 & 74.4 \\
dogs02 & 91.6 \\
farm01 & 82.6 \\
giraffes01 & 65.9 \\
goats01 & 89.8 \\
horses02 & 86.2 \\
horses04 & 88.6 \\
horses05 & 71.6 \\
lion01 & 84.9 \\
marple12 & 79.3 \\
marple2 & 73.7 \\
marple4 & 87.8 \\
marple6 & 50.8 \\
marple7 & 32.1 \\
marple9 & 38.4 \\
people03 & 42.9 \\
people1 & 86.1 \\
people2 & 88.0 \\
rabbits02 & 93.8 \\
rabbits03 & 85.9 \\
rabbits04 & 20.2 \\
tennis & 78.6 \\
\hline
Frame Avg & 72.4 \\
\shline
\end{tabular}
\caption{\textbf{Per sequence Jaccard index $\mathcal{J}$ on FBMS59 \cite{brox2010freiburg,ochs2013segmentation}.}}
\label{tab:fbms59_per_seq}
\end{table}
}

\def\tabResidualFlowInitUpperBound#1{
\begin{table}[#1]
\setlength{\tabcolsep}{2.25pt}
\centering
\begin{tabular}{lcccccccc}
\shline
Upper bound $\lambda$ & 1 & 5 & \textbf{10} & 20 & 50 & 100 & 200 & 400 \\
\shline
\textbf{Ours Init} & 72.7 & 76.5 & \textbf{78.9} & 78.3 & 78.3 & 77.4 & 72.8 & 78.3 \\
Default Init & 72.7 & 76.0 & 78.1 & 78.5 & 73.5 & 73.4 & 73.3 & {\color{red}1.0} \\
\shline
\end{tabular}
\caption{\textbf{Using a small initialization and upper bound is important for the residual flow pathway in our method.} Ours Init refers to an initialization scheme which is 10x smaller than PyTorch default init. Red color indicates {\color{red}collapses}.}
\label{tab:residual-flow-init-upper-bound}
\end{table}
}

\def\tabMotionAppearanceWD#1{
\begin{table}[#1]
\setlength{\tabcolsep}{8pt}
\centering
\begin{tabular}{lccccc}
\shline
Weight Decay & $10^{-6}$ & $10^{-4}$ & $10^{-2}$ \\
\shline
Motion-app. Alignment & -0.672 & \textbf{-0.670} & -0.768 \\
Subset 1 mIoU & 77.2 & \textbf{77.6} & 75.7 \\
Subset 2 mIoU & 77.0 & \textbf{80.5} & 72.0 \\
Subset 3 mIoU & \textbf{77.3} & 76.8 & 76.2 \\
Full val mIoU & 77.2 & \textbf{78.9} & 74.8 \\
\shline
\end{tabular}
\caption{\textbf{Applying motion-appearance alignment provides the optimal weight decay without using labels.} In contrast, using subset mIoU misses the optimal value in one of the three runs. Higher metric values indicate higher segmentation quality for all metrics.}
\label{tab:motion-appearance-alignment-wd}
\end{table}
}

\def\algHyperparamTuning#1{
\begin{algorithm}[#1]
\caption{Pseudo-code for using motion-appearance alignment for hyperparameter tuning}
\label{alg:hyperparameter-tuning}
\begin{algorithmic}
\Require A set of frames $\{I\}$ with $N$ frames
\Require A set of settings with different hyperparameter values $\{S\}$
\Ensure A chosen optimal setting $S^*$ according to motion-appearance-alignment
\For{each setting $S$ in $\{S\}$}
\State Train a model with setting $S$
\State Obtain prediction masks $\{M\}$ with trained model
\For{each frame-mask pair ($I_i$, $M_i$) in $\{I\}, \{M\}$}
    \State Calculate affinity $A$ from frozen ViT features:
    \State $A_{ij}=\mathbb{1}(\text{sim}(f_\text{aux}(I_t)_i, f_\text{aux}(I_t)_j) \geq 0.2)$
    \State Calculate cut between the predicted foreground and background $\text{Cut}(A, \vec{x})$:
    \State $\vec{x} \gets \text{Flatten}(M_i)$
    \State $\text{Cut}(A, \vec{x}) = (1-\vec{x}) A \vec{x}$
    \State Calculate normalized cut between the predicted foreground and background $\text{NCut}(A, \vec{x})$:
    \State $\text{NCut}(A, \vec{x}) = \frac{\text{Cut}(A, \vec{x})}{\sum_{i=1}^{HW} (A\vec{x})_i}
        + \frac{\text{Cut}(A, \vec{x})}{\sum_{i=1}^{HW} (A(1-\vec{x}))_i}$
    \State Calculate the motion-appearance alignment for the current frame:
    \State $L_i \gets -\text{NCut}(A, \vec{x})$
\EndFor
\State $L_{S} \gets \frac{1}{N}\sum_{i=1}^N L_i$
\EndFor
\State $S^* = \arg \max_{S} L_S$
\end{algorithmic}
\end{algorithm}
}

\figVisualizationsSuppResidual{!p}
\figVisualizationsSuppDAVIS{!p}
\figVisualizationsSuppST{!p}
\figVisualizationsSuppFBMS{!p}

\section{Additional Visualizations and Discussions}
We present additional visualizations on the three main datasets that we benchmark our method on \cite{perazzi2016benchmark,li2013video,brox2010freiburg,ochs2013segmentation}. We demonstrate high-quality segmentation in several challenging cases and discuss some limitations of our method with examples.

\subsection{Visualizations of the Residual Pathway}
As shown in \cref{fig:visualizations_residual}, the introduction of the residual pathway allows our segmentation prediction to better fit the flow of deformable and articulated objects. In addition, it also relieves our segmentation module from strictly fitting the flow from 3D rotation and changing depth in a piecewise constant manner. By modeling relative motion in 2D flow, the residual pathway makes our method flexible and robust to objects with complex motion.

\subsection{DAVIS2016, SegTrackv2, and FBMS59}
We visualize our methods on DAVIS2016, SegTrackv2, and FBMS59 in \cref{fig:visualizations_davis}, \cref{fig:visualizations_st}, and \cref{fig:visualizations_fbms59}, respectively. Our method shows great robustness in challenging scenes where there is insufficient motion information, due to its ability to leverage both motion and appearance.

\section{Additional Experiments}
Unless otherwise stated, all the ablation experiments in this section include only stage 1, as the ablations in this section are not relevant to the appearance supervision. Results are without post-processing.

\subsection{Abltion on Different Optical Flow Estimation Methods}
As listed in \cref{tab:flow-method}, almost all recent UVOS works rely on a separate optical flow model pretrained on synthetic data. We use RAFT \cite{teed2020raft} flow by default, following previous works in UVOS. AMD trains \cite{sun2017pwc} from scratch but achieves much lower mIoU.

To evaluate our method's robustness to optical flow estimation methods, we evaluate our method on PWCNet \cite{sun2017pwc}, GMFlow \cite{xu2022gmflow}, and self-supervised ARFlow \cite{liu2020learning}, in addition to RAFT \cite{teed2020raft}.

As shown in \cref{tab:ablation-flow-method}, our method suffers from a mild drop with noisier optical flow. However, our performance is largely retained without tuning the hyperparameters when employing other optical flow methods. We believe the performance gap between different optical flow estimation methods will be reduced further with additional hyperparameter tuning on each flow estimation method.

\tabFlowMethod{t!}
\tabAblationFlowMethod{t!}

\subsection{Preventing Trivial Solutions for Residual Flow Prediction}
There are two factors that prevent trivial solutions: \textbf{1)} Regularization with upper bound $\lambda$ limits the residual prediction to only capturing small relative motion~(10 pixels by default). \textbf{2)} The residual flow branch is initialized to be small, which favors the solution to be simple motion patterns.

As shown in \cref{tab:residual-flow-init-upper-bound}, the results (mIoU on DAVIS16) show that small residual initialization allows RCF to be insensitive to a large range of $\lambda$ against performance degradation or {\color{red}collapses}, even though setting $\lambda$ too large will still cause instability in the form of large mIoU fluctuations. With small residual initialization, $\lambda$ is relatively stable to tune. 

\subsection{Applying Motion-appearance Alignment to Non-method Specific Hyperparameters}
To explore the possibility of using our proposed label-free hyperparameter tuning method to tune hyperparameters that are non-method specific, we evaluate our metric on runs with three different weight decay values: $10^{-6}$ and $10^{-2}$ in addition to our default value of $10^{-4}$. We choose this range of hyperparameter values since we observed that varying the weight decay by smaller amounts had a negligible impact on the final mIoU. As in other hyperparameter tuning experiments, we randomly sample $25\%$ of the sequences from the validation set three times and evaluate the effect of using a smaller labeled validation subset for comparison. Shown in \cref{tab:motion-appearance-alignment-wd}, while the mIoU values from the labeled validation subsets vary significantly between samplings, with one of the three runs missing the optimal value, our metric follows the full validation mIoU trend and selects the best hyperparameter values among the three.

\tabResidualFlowInitUpperBound{t}
\tabMotionAppearanceWD{t}

\section{Pseudo-code for Hyperparameter Tuning With Motion-appearance Alignment}
\algHyperparamTuning{t}
We present the pseudo-code for hyperparameter tuning with motion-appearance alignment in \cref{alg:hyperparameter-tuning}.

\section{Additional Implementation Details}

Our setting mostly follows previous works \cite{liu2021emergence,choudhury2022guess}. Following the official implementation in \cite{liu2021emergence}, we treat the video frame pair $\{t, t+1\}$ as both a forward action from time $t$ to time $t+1$ and a backward action from time $t+1$ and $t$, since they follow similar rules for visual grouping. Therefore, we use this to implement a symmetric loss that applies the loss function on both forward and backward. We then sum the forward loss and backward loss up to obtain the final loss. Note that this could be understood as a data augmentation technique that always supplies a pair in forward and backward to the training batch. However, since our ResNet shares weights for each image input, the feature for each input is reused by the forward and backward action. Furthermore, our residual prediction head has four times the number of channels of the segmentation head to separately predict the forward/backward flow in horizontal/vertical directions, due to its better performance. Thus, the symmetric loss only adds marginal computation and is included in our implementation as well.

Furthermore, following \cite{liu2021emergence}, for DAVIS16, we use random crop augmentation during training to crop a square image from the original image. At test time, we directly input the original image, which is non-square. It is worth noting that the augmentation makes the image size different for training and testing, but as ResNet \cite{he2016deep} takes images of different sizes, this does not pose a problem empirically. In STv2 and FBMS59, the images have very different aspect ratios (some having a height lower than the width), and thus we resize the images to 480p as a preprocessing before the standard pipeline. We additionally use pixel-wise photo-metric transformation \cite{mmseg2020} for augmentation with the default hyperparameters for this augmentation.

As for the architecture, we found that simply taking the feature from the last ResNet stage provides insufficient detailed information for high-quality output. Instead of incorporating a more complicated segmentation head (\eg, \cite{cheng2021maskformer} in \cite{choudhury2022guess}), we chose to keep our architecture easy to implement by only changing the head in a simple fashion. Following the standard approach of multi-scale feature fusion, we resized and concatenated the feature from the first residual block and the last residual block in ResNet, which allows the feature to jointly capture high-level information and low-level details. Note that such fusion is only applied to the segmentation head, and residual prediction is simply bilinearly upsampled. Due to lower image resolution, no feature merging is performed for STv2 in stage 1. Following \cite{choudhury2022guess}, we load self-supervised ImageNet pretrained weights learned without annotation, since the training video datasets are too small for learning generalizable feature (\eg, DAVIS16/STv2/FBMS59 has only 3,455/976/13,860 frames), with DenseCL weights \cite{ILSVRC15,wang2021dense} on ResNet50 for our method. This can be replaced by training on uncurated Youtube-VOS \cite{xu2018youtube} with our training process, as in \cite{liu2021emergence}, so that one implementation can be used throughout training for simplicity in real-world applications.

In our training, we follow \cite{liu2021emergence} and use a batch size of 16 (with two images in a pair, and thus 32 images processed in each forward pass). Stage 1 and stage 2 take 200 and 40 epochs, respectively, for DAVIS16. We use a learning rate of $1\times10^{-4}$ with Adam optimizer \cite{kingma2014adam} and polynomial decay (factor of $0.9$, min learning rate of $1\times10^{-6}$). We set weight decay to $1\times10^{-4}$ for DAVIS and $1\times10^{-6}$ for STv2 and FBMS59. Due to the fact that normalized cuts is slow to optimize, we split stage 2 into two sub-stages: one with the CRF followed by one with normalized cuts optimization, each of the stage has the same number of training steps. In the CRF substage in stage 2, we set $w_\text{motion}=1$ and $w_\text{app}=10$ to balance the two losses. However, we observe training instability if we supervise the network directly by its output refined by the CRF. Therefore, we apply exponential moving averaging (EMA) to the model weights and supervise the network by the output from the EMA model, with momentum $m=0.999$. In the normalized cuts substage, we pre-generate the network's outputs and use the refinement as described in the methods section, which involves running CRF before and after normalized cuts refinement and multiplying the refined masks from the two CRF runs. This is equivalent to applying such refinement with EMA with $m=1.0$. In this substage, we set $w_\text{motion}=0.1$ and $w_\text{app}=2.0$.

\section{Per-sequence Results}
\tabDAVISPerSequence{p}
\tabSTPerSequence{p}
\tabFBMSPerSequence{p}

We list our per-sequence results on DAVIS16 \cite{perazzi2016benchmark}, STv2\cite{li2013video}, FBMS59\cite{brox2010freiburg,ochs2013segmentation} in \cref{tab:davis16_per_seq}, \cref{tab:stv2_per_seq}, and \cref{tab:fbms59_per_seq}, respectively. The results are with post-processing.

\section{Future Directions}
As our method does not impose temporal consistency, it does not effectively leverage information redundancy from neighboring frames. Using such information could make our method more robust in dealing with frames that provide insufficient motion and appearance information. Temporal consistency measures, such as matching warped predictions, could be incorporated as an additional loss term or as post-processing, as in \cite{yang2019unsupervised}.

Furthermore, our method currently does not support segmenting multiple parts of the foreground or identifying each object instance. To address this, methods such as normalized cuts \cite{shi2000normalized} could be used to split the foreground into several objects with motion and appearance input to provide signals to train the model. Another potential approach is to over-split the scene with many object channels and use other unsupervised methods such as FreeSOLO \cite{wang2021dense, wang2022freesolo} to obtain coarse segmentation proposals to merge the channels to form object instance segmentation.

\FloatBarrier

\clearpage

{\small
\bibliographystyle{ieee_fullname}
\bibliography{egbib}
}

\end{document}